%% file: sdm.tex
\documentclass[11pt]{article}

\usepackage[preprint]{acl}  

\usepackage{times}
\usepackage{latexsym}

\usepackage[T1]{fontenc}

\usepackage[utf8]{inputenc}

\usepackage{microtype}

\usepackage{inconsolata}

\usepackage{graphicx}

\usepackage[table]{xcolor}         

\usepackage{booktabs}       
\usepackage{amsmath}
\usepackage{amssymb}
\usepackage{mathtools}
\usepackage{amsthm}
\usepackage{basecommon} 

\colorlet{blueHighlightColor}{blue!05} 
\colorlet{wrongPredictionColor}{red!05} 
\colorlet{correctPredictionColor}{green!05} 
\definecolor{lightestgray}{gray}{0.95}
\colorlet{blueEmphasis}{blue!80}

\usepackage{caption}
\usepackage{listings}
\theoremstyle{plain}
\newtheorem{theorem}{Theorem}[section]

\theoremstyle{definition}

\newtheorem{assumption}[theorem]{Assumption}
\theoremstyle{remark}

\usepackage{algorithm}
\usepackage[noend]{algpseudocode}  
\algrenewcommand\alglinenumber[1]{\small #1:}

\title{Similarity-Distance-Magnitude Activations}

\author{Allen Schmaltz \\
  Reexpress AI \\
  \texttt{allen@re.express} \\}

\begin{document}

\maketitle

\begin{abstract}
We introduce the \textsc{Similarity}-\textsc{Distance}-\textsc{Magnitude} (\textsc{sdm}) activation function, a more robust and interpretable formulation of the standard $\softmax$ activation function, adding \textsc{Similarity} (i.e., correctly predicted depth-matches into training) awareness and \textsc{Distance}-to-training-distribution awareness to the existing output \textsc{Magnitude} (i.e., decision-boundary) awareness, and enabling interpretability-by-exemplar via dense matching. We further introduce the $\sdm$ estimator, based on a data-driven partitioning of the class-wise empirical CDFs via the $\sdm$ activation, to control the class- and prediction-conditional accuracy among selective classifications. When used as the final-layer activation over pre-trained language models for selective classification, the $\sdm$ estimator is more robust to covariate shifts and out-of-distribution inputs than existing calibration methods using $\softmax$ activations, while remaining informative over in-distribution data.
\end{abstract}

\section{Introduction}

Neural-network-based language models (LMs) pose a challenge for interpretable and reliable deployment given the non-identifiability of their parameters \citep[inter alia]{hwang-and-ding-1997}\footnote{Informally, this means that two or more distinct sets of values for the parameters can result in identical output distributions. As a consequence, interpreting the parameters of such models is typically more complicated than with a simple linear regression model, for example.}, which can number in the billions or more. Instead of directly interpreting parameters, one option is to move the focus of interpretation to auditing predictions as a form of interpretability by example, or \textit{exemplar}, over the representation space of such models via dense matching \citep{schmaltz-2021-insights}. However, for real-world deployments, robust approaches for predictive uncertainty are also needed, both for human decision-making and for constructing sequentially dependent LM pipelines.

Known theoretical results limit the statistical quantities that can be derived over LMs. Statistical guarantees in the distribution-free setting are limited to approximately conditional quantities \citep[inter alia]{Valiant-1984-PAC,LeiAndWasserman-2014-PredictionBands,BarberEtAl-2020-LimitsOfDistributionFree}. Further, even typical approximately conditional quantities can be difficult to obtain in practice, since the minimal assumption of exchangeability with a known held-out dataset is itself often violated with covariate and label shifts, which can be difficult to foresee with existing methods. Epistemologically, the prevalence of hallucinations and highly-confident wrong answers with widely deployed LMs suggests a technical impasse in effectively modeling the predictive uncertainty, despite significant work from Bayesian, Frequentist, and empirically motivated perspectives \citep[inter alia]{GalAndZoubin-2016-MCDropout,Angelopoulos-2021-RAPS,GuoEtAl-2017-TempScaling,Lakshminarayanan-2017-DeepEnsembles,Ovadia-EtAl-2019-EvaluatingUncertainty}. A foundational piece is evidently missing from the picture.

Given these intrinsic challenges, we approach the problem of uncertainty quantification over LMs from a new angle and ask: \textit{Can we leverage the metric learning and dense matching capabilities of neural networks over high-dimensional inputs to at least aim to decompose signals of epistemic (reducible) uncertainty in a manner that is interpretable and actionable?}

We address this question with a conceptually parsimonious, data-driven partitioning of the data to decompose sources of epistemic uncertainty: Correctly predicted depth-matches into the training set ($\Similarity$), the $\Distance$ to the training set, and the distance to the decision-boundary ($\Magnitude$). We use these signals to construct a new activation function, the $\sdm$ activation, which can be used as a replacement for the standard $\softmax$ activation, as, for example, the final-layer activation. The $\sdm$ activation enables more reliable estimates of the predictive uncertainty for selective classification \citep[inter alia]{Chow-1957-EarlyPredictWithRejectSystem,GeifmanAndEl-Yaniv-2017-NN-SelectiveClassification}, which addresses the need for uncertainty quantification with LMs used in multi-stage decision pipelines, in settings subject to covariate shifts and out-of-distribution inputs. 

In summary, in this work:
\begin{itemize}
\item We introduce the $\Similarity$-$\Distance$-$\Magnitude$ ($\sdm$) activation function. 
\item We introduce the $\sdm$ estimator for use in controlling class- and prediction-conditional accuracy among selective classifications, based on a data-driven partitioning of the class-wise empirical cumulative distribution functions (eCDFs) over the $\sdm$ activation output.  
\item We examine the behavior of the $\sdm$ activation as a final-layer activation over pre-trained language models, using the $\sdm$ estimator for selective classification. We demonstrate empirically that the $\sdm$ estimator is more robust to covariate-shifts and out-of-distribution inputs than existing classes of post-hoc calibration methods, while remaining informative over in-distribution data.
\end{itemize}
\section{Preliminaries}

\subsection{Setting}
We consider the standard multi-class classification setting. We are given a training dataset, $\trainSplit=\{(\vx_n, y_n)\}_{n=1}^{N}$ of inputs, $\vx \in \gX$, paired with their corresponding ground-truth discrete labels, $y \in \gY = \{1, \ldots, C\}$, and a labeled calibration dataset, $\calibrationSplit$, drawn from the same distribution as $\trainSplit$. We are then given a new test instance, $\vx$, from an unlabeled test set, $\testSplit$, and seek to estimate the label with a prediction, $\hat{y}$, via the un-normalized log probabilities (``logits'', informally) of a final linear layer: $\vz  = \mW^T \vh + \vb$, where $\vh = \underlyingNetwork(\vx; \theta)$ is the final hidden state of a network parameterized by $\theta$. The discrete prediction is taken as $\hat{y} = \argmax{\vz}$; however, for learning $\theta$, $\mW$, and $\vb$, and for human decision-making, we also seek an estimate of the predictive uncertainty, $p(y \given \vx)$, which is typically obtained by normalizing $\vz$ via the $\softmax$ activation described next. We will make a distinction between models, $\gM$ (defined by $\theta$, $\mW$, and $\vb$, and when applicable, the exemplar adaptor, described below), which produce the prediction, $\hat{y}$, and estimators, $\gE$, which provide an estimate of $p(y \given \vx)$, because different estimators can be used over the same model. 

\subsection{Softmax and the Cross-Entropy loss}
 
The $\softmax$ activation is commonly used in neural network architectures, including, for example, as a router in self-attention mechanisms \citep{VaswaniEtAl-2017-AttentionIsAllYouNeed} and mixture-of-experts models \citep{ShazeerEtAl-2017-MOE}, and forming the basis of the cross-entropy loss used for next-token training of LMs. It is the typical final output layer of LMs, converting the un-normalized model logits to a normalized probability distribution: 
\begin{align}\label{eq:softmax}
\softmax(\vz)_i = \frac{
e^{\tau \cdot z_i}
}{
\sum^C_{c=1}{e^{\tau \cdot z_c}}
}, 1 \le i \le C, \tau \ge 0
\end{align}
The inverse-temperature parameter, $\tau$, controls the sharpness of the distribution. As $\tau \rightarrow 0$, the output of $\softmax(\vz)$ converges to a uniform distribution where each class has probability $\frac{1}{C}$; as $\tau \rightarrow \infty$, the output converges to a distribution in which all of the mass is assigned to a single class. In deep learning, $\tau$ is treated as a learnable, \textit{global} hyper-parameter; \textit{instance-wise} variation in the distance to the decision-boundary is thus determined by the relative $\Magnitude$ of $z_{\hat{y}}$. This model is learned by minimizing the cross-entropy loss between $\vz$ and the index of the true labels over $\trainSplit$. The \textit{natural} logarithm of the loss is the counterpart to the base $e$ of the $\softmax$:
\begin{align}\label{eq:cross-entropy-loss}
\gL(\theta, \mW, \vb; \trainSplit) = -\frac{1}{N} \sum_{n}^{N} \log_{e}\left(
\frac{
e^{\tau \cdot z_{y_n}}
}{
\sum^C_{c=1}{e^{\tau \cdot z_c}}
}
\right)
\end{align}
\section{Methods}
In this work, we revisit Eq.~\ref{eq:softmax} and ~\ref{eq:cross-entropy-loss}. We seek to decouple the sources of epistemic uncertainty via a new activation function that is conceptually:
\begin{align}\label{eq:abstractSDM}
\sdm(\vz)_i = 
\frac{
{\Similarity}^{\Distance \cdot \Magnitude_i}
}{
\sum^C_{c=1}{{\Similarity}^{\Distance \cdot \Magnitude_c}}
}
\end{align}
with a corresponding negative log likelihood loss that takes into account the change of base (\S~\ref{sec:sdm-activation}). Unique to this setting, a modification to label-conditional conformal prediction \citep{Vovk-2005-AlgorithmicLearningBook} then follows via a parsimonious partitioning of the class-wise empirical CDFs, providing a principled basis for controlling the class-conditional accuracy among selective classifications, combined with empirically-robust prediction-conditional estimates.

\subsection{Similarity-Distance-Magnitude Activation Functions}\label{sec:sdm-activation}

Calculating the $\sdm$ activation involves training an exemplar adaptor, a 1-D CNN adaptor (with a final linear layer) over the frozen hidden states of a network, to induce distilled, compressed representations of the underlying network's representation space conditional on its predictions. The resulting representations provide a probabilistic mapping to the training, or support, set. In this way, neural networks, including large pre-trained networks, can be viewed as \textit{hidden} instance-based metric learners \citep{schmaltz-2021-insights}, from which we can then derive signals of the epistemic uncertainty.

\subsubsection{Exemplar Adaptor}
We take as the CNN of our exemplar adaptor $g: \vh \in \reals^D \mapsto \vh' \in \reals^M$, a 1-D CNN that takes as input $\vh$ of the underlying network. The CNN has $M$ filters, the filter applications of which produce $\vh'$, the distilled representation of the underlying network. A final linear layer, $\vz'  = \mW'^T \vh' + \vb', \vz' \in \reals^C$, then replaces the underlying network's final linear layer, with the discrete prediction taken as $\hat{y} = \argmax{\vz'}$. This exemplar adaptor will then enable us to derive the $\Similarity$, $\Distance$, and $\Magnitude$ values, as defined next.

\subsubsection{$\Similarity$}\label{sec:similarity-calculation} 

We define the $\Similarity$ ($\q$) of an instance to the training set as the count of consecutive nearest matches in $\trainSplit$ that are correctly predicted \textit{and} match $\hat{y}$ of the test instance.\footnote{We use the letter $\q$, as this value \textbf{q}uantizes the closeness of a point to the training set with a discrete estimate.} Concretely, we first sort $\gD_{\rm{tr}}$ (for which we have both model predictions and ground-truth labels) based on the $L^2$ distance from $\vh'$, $\left[(\vx^{tr}_{(1)}, \hat{y}^{tr}_{(1)}, y^{tr}_{(1)}),\ldots, (\vx^{tr}_{(N)}, \hat{y}^{tr}_{(N)}, y^{tr}_{(N)})\right]$, such that $|| \vh' - \vh'^{tr}_{(1)} ||_2 \le \ldots \le || \vh' - \vh'^{tr}_{(N)} ||_2$, and then calculate $q \in \{0, \ldots, | \trainSplit |\}$ as:
\begin{align}\label{equation:q} 
&q = \notag \\
&\sum^{\vert \trainSplit \rvert}_{i=1} \mathbf{1}_{\hat{y} =  \hat{y}^{\rm{tr}}_{(i)}} \cdot \mathbf{1}_{\hat{y}^{\rm{tr}}_{ (i) } = y^{\rm{tr}}_{ (i) }} \cdot \mathbf{1}_{
i-1 = \sum^{i-1}_{j=1} \mathbf{1}_{\hat{y} =  \hat{y}^{\rm{tr}}_{(j)}} \cdot \mathbf{1}_{\hat{y}^{\rm{tr}}_{ (j) } = y^{\rm{tr}}_{ (j) }} 
}
\end{align}
where the rightmost indicator function, $\mathbf{1} \in \{ 0, 1\}$, ensures consecutive (depth-wise) matches.\footnote{This seemingly simple rule differs from traditional KNN rules \citep[inter alia]{CoverAndHart-1967-KNNBayesError,DevroyeEtAl-1996-APT} in two critical respects: The neural network serves as a semi-supervised metric learner of the distances between the dense representations that identify the instances, and there is a model prediction (in addition to the ground-truth label) for each instance in the support set. The former enables effective partitioning despite the curse of high dimensions; the latter provides an additional indicator of reliability for each instance.} By definition, $\q$ cannot exceed the count of the most prevalent class label in $\trainSplit$, and since we assume an approximately equal number of points for each class, $\q \ll | \trainSplit |$ is typical. For the special case of calculating $q$ for $\vx \in \trainSplit$, which only occurs during learning, we exclude the self-match.

\subsubsection{$\Distance$}\label{sec:distance-calculation}
The $L^2$ distance to the nearest match in $\trainSplit$ follows from above: $\dNearest = || \vh' - \vh'^{tr}_{(1)} ||_2$. We normalize these values by defining the $\Distance$, $\dVal \in [0,1]$, in terms of the class-wise empirical CDFs of $\dNearest$ over $\calibrationSplit$, as the most conservative quantile relative to the distance to the nearest matches observed in the labeled, held-out set:
\begin{align}\label{equation:d} 
\dVal = \min~ &[ 1-{\rm{eCDF}}^{y_1}_{\rm{ca}}( \dNearest ) , \ldots , \notag \\ &~1-{\rm{eCDF}}^{y_C}_{\rm{ca}}( \dNearest ) ]
\end{align}
The empirical CDFs are determined by the labeled points in $\calibrationSplit$ for which $\q>0$, where, as indicated by the superscripts, the stratification of points is by the true labels, $y$. For example, ${\rm{eCDF}}^{y_1}_{\rm{ca}}( \dNearest )$ is the empirical CDF of $\dNearest$ values in $\calibrationSplit$ for which $y=1$, a notation convention we will use throughout. (Points with $\q=0$ are effectively out-of-distribution points and treated as such in downstream decision-making, so they are excluded to avoid biasing the estimates.) At test time, we do not see $y$; instead, the minimum is calculated over the quantiles of each of the class-conditional eCDFs, regardless of $\hat{y}$. As with $q$, for the special case of calculating $\dVal$ for $\vx \in \trainSplit$, we replace ${\rm{eCDF}}^{y_c}_{\rm{ca}}$ with the analogous ${\rm{eCDF}}^{y_c}_{\rm{tr}}$, the class-wise empirical CDFs of $\dNearest$ over $\trainSplit$ excluding self-matches.

Appendix~\ref{appendix:effective-sample-size} provides a method for accounting for error in the estimation of the eCDFs given the effective sample size.
\subsubsection{$\Magnitude$}
We take as the $\Magnitude$, or distance to the decision boundary, $z'_{\hat{y}}$, as in the standard $\softmax$ case but via $\vz'$ from the linear layer of the exemplar adaptor.

\subsubsection{SDM Activation: Formulation}
We use the above quantities to define the $\sdm$ activation function:
\begin{align}\label{eq:sdmActivation}
\sdm(\vz')_i = \frac{
(2+q)^{\dVal \cdot z'_i}
}{
\sum^C_{c=1}{(2+q)^{\dVal \cdot z'_c}}
}, 1 \le i \le C
\end{align}
The output distribution becomes sharper with larger values of $\q$ and $\dVal$, as well as with larger relative values of $z'_{\hat{y}}$, as with the standard $\softmax$. When $\dNearest$ exceeds the largest distance observed in the labeled data, $\dVal=0$ and the output distribution is uniform, reflecting maximally high uncertainty. The standard $\softmax$ with $\tau=1$ is recovered by setting $q = e-2, d=1$. Provided $\dVal \neq 0$, $\argmax \sdm(\vz') = \argmax \vz'$, as with the $\softmax$ activation.
 
\subsubsection{SDM Activation: Loss and Training}

A loss analogous to Eq.~\ref{eq:cross-entropy-loss} then follows with the applicable change of base. We use this loss to train the weights of the exemplar adaptor, which includes the parameters of the linear layer ($\mW'$ and $\vb'$), as well as the convolution weights and biases, which we collectively represent with $\mG$. The weights of the underlying $\underlyingNetwork$ remain fixed.
\begin{align}\label{eq:sdm-loss}
&\gL(\mG, \mW', \vb'; \trainSplit) = \notag \\ &-\frac{1}{N} \sum_{n}^{N} \log_{(2+q)}\left(
\frac{
(2+q)^{\dVal \cdot z'_{y_n}}
}{
\sum^C_{c=1}{(2+q)^{\dVal \cdot z'_c}}
}
\right)
\end{align}
The first epoch of training is initialized with a standard $\softmax$ (i.e., setting $q = e-2, d=1$). Training then proceeds by re-calculating $q$ and $d$ for each $\vx \in \trainSplit$ after each epoch. We take as the stopping criteria for one learning round as the epoch with the lowest balanced (across classes) average loss over $\calibrationSplit$. We repeat this process for $J$ iterations of random shuffles and splits of $\trainSplit$ and $\calibrationSplit$ and parameter initializations, choosing the final model as that with the globally lowest balanced (across classes) average loss over $\calibrationSplit$. 

\subsection{Evaluating Selective Classification}

As a common, unambiguous baseline quantity for comparing selective classifiers over a held-out test set, $\testSplit$, we seek an easy-to-interpret and easy-to-evaluate metric, reflecting real-world applications. Among the selective classifications from an estimator, we seek (Quantity I) prediction-conditional accuracy at or above a given threshold, $\alpha \in (\frac{1}{2}, 1]$, and (Quantity II) class-conditional accuracy at or above that same threshold, $\alpha$.

To evaluate this metric, we only consider the points for which the given estimator assigns a high-probability of at least $\alpha$, which is typically near 1, such as $\alpha=0.95$ in our experiments. We refer to this set of points as the \textit{admitted}, or \textit{non-rejected}, set. Then, given ground-truth values for $\testSplit$, we assess whether the conditional accuracies of the admitted set are at least $\alpha$ when (Quantity I) stratifying by the predicted labels, $\hat{y}$, and when (Quantity II) stratifying by the true labels, $y$. 

The estimator that rejects all points would meet these conditions. However, given two estimators that meet these conditions, we prefer that which rejects fewer points, ceteris paribus. In other words, we seek estimators that meet our reliability condition and are informative (i.e., maximize the number of points that are properly admitted), but when the estimator is uncertain, we prefer rejection over unexpectedly falling under the desired $\alpha$ probability threshold.

Quantity I corresponds to top-label calibration \citep{GuptaAndRamdas-2022-ToplabelCalibration}, but with a single bin for evaluation, $[\alpha, 1]$, removing ambiguity with regard to the choice of binning the probabilities. Quantity II does not directly correspond to quantities typically examined in the calibration literature \citep[inter alia]{Brier-1950-BrierCalibration,Dawid-1982-CalibratedBayesian,GuoEtAl-2017-TempScaling,VaicenaviciusEtAl-2019-EvaluatingCalibration,KullEtAl-2019-BeyondTempScaling}, but it approximates label-conditional conformal coverage in the special case of class-wise thresholds that only admit prediction sets of cardinality 1. We introduce a straightforward procedure to estimate this quantity next.

\subsubsection{Controlling the Class-conditional Accuracy among Selective Classifications with SDM Estimators}\label{sec:minRescaledSimiliarityForHRRegion}
In general, the statistical coverage guarantee of marginal split-conformal estimators is not directly informative for selection, since the coverage guarantee is not conditional on the set size. We may instead seek one of various approximately conditional notions of coverage \citep[inter alia]{RomanoEtAl-2020-APS,Angelopoulos-2021-RAPS}; however, there is no guarantee that when we stratify by sets of cardinality 1, coverage will be maintained. However, there is a \textit{special case} in which label-conditional conformal estimators \textit{do} provide a meaningful notion of class-conditional coverage for selection. Assuming $\calibrationSplit$ and $\testSplit$ are exchangeable, if the conformity score for each label is from a categorical distribution and the resulting thresholding of the class-wise empirical CDFs results in class-wise thresholds that are all greater than $\frac{1}{2}$, then the cardinality 1 sets will, on average, obtain class-conditional coverage, by definition.\footnote{Although the formal interpretations are not identical, the evaluation of class-conditional coverage for a single estimate of such cardinality 1 prediction sets in this restricted setting over $\testSplit$ is numerically equivalent to assessing the class-conditional accuracy, when not considering the sample-size dependent error term. See Appendix~\ref{appendix:effective-sample-size} for a further discussion of analyzing the effective sample size.} Unfortunately, it may be rare to encounter this restricted setting over the full data distributions of real-world tasks. Instead, we will use the $\sdm$ activation to estimate a partitioning of the distribution into a region that approximately fulfills these assumptions.

First, we rescale the $\Similarity$ estimate to take into account the $\Distance$ and $\Magnitude$, given the predicted class. The resulting value\footnote{The $\min$ ensures that $\rescaledSimilarity$ remains 0 for points with $q=0$, which are effectively out-of-distribution points.} will be the basis for partitioning the distribution:
\begin{equation}\label{equation:rescaled-similarity} 
\rescaledSimilarity = \min \left ( \q, (2+q)^{\sdm(\vz')_{\hat{y}}} \right )
\end{equation}

Next, we estimate label-conditional conformal thresholds, $[\psi_1, \ldots, \psi_C]$, over the output from the $\sdm$ activation among a subset of the distribution constrained by progressively larger values of $\rescaledSimilarity$ (among $\rescaledSimilarity > 0$) until all thresholds are at least $\alpha$. By setting the stopping criteria at $\alpha$ rather than $\frac{1}{2}$, we also restrict the region to the empirically-motivated prediction-conditional quantity, $\sdm(\vz')_{\hat{y}}$. The procedure appears in Alg.~\ref{alg:estimate-high-reliability-region}. If we find a finite $\minRescaledSimiliarityForHRRegion$ that obtains such thresholds, we refer to the resulting region as the $\hrRegionFull$ ($\sdmHR$) region, taking membership in this region as our selection criteria:
\begin{align}\label{eq:index-conditional-sdm-estimator}
\sdmHR := 
\begin{cases}
  \hat{y} & \text{if~}  \rescaledSimilarity \ge \minRescaledSimiliarityForHRRegion  \wedge \sdm(\vz')_{\hat{y}} \ge \psi_{\hat{y}}  \\
  \bot & \text{otherwise}
\end{cases}
\end{align}
where $\bot$ indicates a rejected (non-admitted) point and $\hat{y} = \argmax{\vz'}$.\footnote{The convention is to refer to the basic architecture of Eq.~\ref{eq:sdmActivation} as the $\sdm$ activation function, and using the activation with the selection criteria of Eq.~\ref{eq:index-conditional-sdm-estimator} as the $\sdm$ estimator.}

To calculate this quantity for new, unseen test points $\vx \in \testSplit$, we require $\trainSplit$ to calculate $\q$ and $\dNearest$; the cached class-wise empirical CDFs of the distances over $\calibrationSplit$ of Eq.~\ref{equation:d}; and $\minRescaledSimiliarityForHRRegion$ and the thresholds, $[\psi_1, \ldots, \psi_C]$. Evaluation of the $\sdmHR$ selection criteria is straightforward: We simply assess the conditional accuracies for the admitted points after stratifying by the predictions and the true labels, each in turn.
\input{algorithms/algorithm-estimate-high-reliability-region.tex}
When Alg.~\ref{alg:estimate-high-reliability-region} returns $\minRescaledSimiliarityForHRRegion=\infty$, we obtain a useful empirical indicator that the model is too weak, or the data insufficient, to reliably obtain class- and prediction-conditional estimates at the specified $\alpha$ value.

\section{Experiments}
We provide controlled comparisons of our proposed methods over representative LMs and tasks, systematically ablating relevant components, holding the data and underlying LM constant, ceteris paribus. We consider in-distribution, covariate shifted, and out-of-distribution test sets. We consider representative estimators over the existing LM architecture (i.e., without additional parameters); with CNN adaptors; and with the $\sdm$ activation layer. Additional details are provided in the Appendix.

\subsection{Task: Binary Sentiment Classification}  
\paragraph{$\datasetSentiment$: $\trainSplit$ and $\calibrationSplit$.}
Our first task is predicting the sentiment of movie reviews. We use the commonly used benchmark data of \citet{Maas-EtAl-2011-OriginalCitedSourceForIMDbReviewsData}. This is a binary classification task with $y \in \{0 = {\rm{negative}}, 1 = {\rm{positive}}\}$. $\trainSplit$ and $\calibrationSplit$ are constructed from a total of 18k instances. The held-out set for evaluation ($\datasetSentiment$), $|\testSplit|=1583$, is from the same distribution as $\trainSplit$ and $\calibrationSplit$. 

\paragraph{$\datasetSentimentOOD$.} To evaluate the behavior of the estimators over out-of-distribution (OOD) data, we consider an additional evaluation set, $\datasetSentimentOOD$, $|\testSplit|=4750$. We use the SemEval-2017 Task 4a test set \citep{Rosenthal-etal-2017-Semeval-Task4}, which consists of short-form social media posts that differ in the distribution of topics, language styles, and lengths relative to the movie reviews. We balance the test set, dropping the third class (neutral), setting the semantics of the true labels to be the same as that of the movie reviews: $y \in \{0 = {\rm{negative}}, 1 = {\rm{positive}}\}$.

\paragraph{Far OOD Challenge Sets.}
In Appendix~\ref{sec:far-ood}, we consider two additional out-of-distribution challenge test sets, $\datasetSentimentShuffled$ and $\datasetSentimentOODShuffled$, constructed by randomly shuffling the input documents for each of $\datasetSentiment$ and $\datasetSentimentOOD$, respectively.

\subsection{Task: $\datasetFactcheck$} 

\paragraph{$\datasetFactcheck$.} As a more challenging binary classification task for LMs, we consider the fact check data of \citet{azaria-mitchell-2023-internal}. The training and calibration sets, a combined total of 6k instances, consist of single sentence statements that have been semi-automatically generated via templates and a knowledge base. The task is to determine whether the statement is true or false,  $y \in \{0 = {\rm{false}}, 1 = {\rm{true}}\}$. The held-out eval set ($\datasetFactcheck$), $|\testSplit|=245$, the focus of our analysis, has been constructed by having an LM generate a statement continued from a true statement not otherwise in the dataset. These evaluation statements are checked manually and assigned labels by human annotators. In addition to being a relatively challenging task that evaluates---at least in principle---the latent knowledge stored within an LM's parameters, the test set is representative of the types of covariate shifts over high-dimensional inputs that can be problematic for real applications, and challenging to characterize without model assistance and ground-truth labels. It was observed in \citet{azaria-mitchell-2023-internal} that the accuracy of existing LM classifiers is lower on this generated, held-out test set compared to the calibration set. However, these test sentences would seem to also be simple true-false statements, reflecting that it is not necessarily straightforward for a human user to detect distribution shifts over high-dimensional inputs. As such, we seek for our models and estimators to reflect such shifts via the predictive uncertainty. 

\paragraph{$\datasetFactcheckShuffled$.} As with the sentiment task, in Appendix~\ref{sec:far-ood} we also consider an additional out-of-distribution challenge test set, $\datasetFactcheckShuffled$, constructed by randomly shuffling the input documents of the $\datasetFactcheck$ task. 

\subsection{Models}\label{sec:models}

We consider two representative, open-weight decoder-only Transformer-based LMs, the parameters of which stay fixed: The 3.8 billion-parameter \texttt{Phi-3.5-mini-instruct} model ($\modelPhiThreeFiveInstruct$) \citep{Abdin-2024-Phi3-TechReport}, and the 47 billion-parameter \texttt{Mixtral 8x7B Instruct v0.1} mixture-of-experts model ($\modelMixtral$) \citep{JiangEtAl-2024-Mixtral8x7B}.

\paragraph{Hidden states.} We take as $\vh$ the concatenation of the final-layer hidden state of the final sequence position (i.e., the hidden state that is the input to the linear-layer over the output vocabulary for the \texttt{Yes} or \texttt{No} generation) with the mean over all final-layer hidden states. For $\modelPhiThreeFiveInstruct$, this results in $\vh \in \reals^{6144}$, and for $\modelMixtral$, $\vh \in \reals^{8192}$.

\subsection{Estimators}

We examine representative Frequentist, Bayesian, and empirically-motivated classes of estimators used with neural networks, setting $\alpha=0.95$ for all experiments. At the most basic, but perhaps the most commonly used in practice, representing the absence of a post-hoc calibration method, we simply threshold the output, $\softmax(\vz) \ge \alpha$, where the temperature $\tau=1$. For this, we use the label $\softmax$. As an established empirical approach for calibrating neural networks, we provide a comparison to temperature scaling \citep{GuoEtAl-2017-TempScaling}, a single parameter version of post-hoc Platt-scaling \citep{Platt-1999-PlattScaling}, with the label $\estimatorTempScaling$. In this case, the estimator is the thresholding of the output $\softmax(\vz; \tau) \ge \alpha$ after learning a value for $\tau$ over $\calibrationSplit$. We also provide a comparison to two representative conformal predictors, the $\conformalAPS$ method of \citet{RomanoEtAl-2020-APS} and the adaptiveness-optimized $\conformalRAPS$ algorithm of \citet{Angelopoulos-2021-RAPS}. The admission criteria for the $\conformalAPS$ and $\conformalRAPS$ estimators is prediction sets of size 1, at the $0.05$ level (i.e., $1-\alpha$, as defined here). We consider these estimators over the logits corresponding to the \texttt{Yes} and \texttt{No} indexes of the output linear-layer of the underlying LM ($\modelPhiThreeFiveInstruct$ and $\modelMixtral$), which provides a reference point without introducing additional adaptor layers. We also consider these baselines over 1-D CNN adaptors over $\vh$ of each LM ($\modelPhiThreeFiveInstructCNNAdaptor$ and $\modelMixtralCNNAdaptor$). These baseline adaptors are identical to those used in the corresponding $\sdm$ activation layers, with $M=1000$, and are similarly trained for $J=10$ iterations of 200 epochs, but unlike the $\sdm$ activation layers, the stopping criteria is the minimum balanced (across classes) average cross-entropy loss.

We further compare to variational Bayesian last-layer neural networks ($\estimatorVBLL$), a computationally efficient Bayesian approach \citep{HarrisonEtAl-2024-VBLL}. We consider the discriminative versions over \texttt{Phi-3.5-mini-instruct} and \texttt{Mixtral 8x7B Instruct v0.1} ($\modelPhiThreeFiveInstructDiscVBLLMLP$ and $\modelMixtralDiscVBLLMLP$, respectively) in the main text, with additional comparisons in Appendix~\ref{appendix:vbll-models}.

We then compare to the final-layer $\sdm$ activations over $\vh$ of each LM ($\modelPhiThreeFiveInstructSDM$ and $\modelMixtralSDM$). For reference, we provide the result of thresholding a $\softmax$ over these adaptors at $\alpha$, as above, as well as a thresholded $\softmax$ that simply treats $\dVal$ as the inverse-temperature, $\estimatorSoftmaxOverDistanceMagnitude$, which is equivalent to setting $q=e-2$ in the $\sdm$ activation. We consider an analogous threshold over the activation output, $\sdm(\vz') \ge \alpha$, for which we use $\sdmAtAlpha$ as the estimator label. Finally, for the eponymous ``$\sdm$ estimator'' using the selection criteria of Eq.~\ref{eq:index-conditional-sdm-estimator}, we use the label $\sdmHR$ in the results tables. 

As a common point of reference, the label $\estimatorNoReject$ refers to predictions without any selective filtering (i.e., the raw output accuracies, derived from the $\argmax$ over the final linear-layer). 

\section{Results}\label{sec:results}

\paragraph{In-distribution data.} Representative results are provided in Table~\ref{tab:experiments-main-text-abbreviated}, with additional results in the Appendix. Even on the in-distribution $\datasetSentiment$ dataset, the estimators over the underlying LMs without adaptor layers exhibit over-confidence, which is reflected in conditional accuracies that fall below the expected $\alpha$. The estimators over the adaptor layers all obtain the desired conditional accuracies, with the class-wise accuracies of the models themselves $\ge \alpha$ (see the $\estimatorNoReject$ rows in Table~\ref{tab:experiments-sentiment}), with differences arising in the proportion of admitted points. Here and elsewhere, $\estimatorSoftmaxOverDistanceMagnitude$ tends to be overly conservative in rejecting points. To be expected, the $\sdmHR$ estimator tends to be more conservative than simply thresholding the $\sdm$ activation at $\alpha$ ($\sdmAtAlpha$), but the latter lacks the assurances on the class-conditional accuracy obtained by the constraints on the $\hrRegionFull$ region. In practice, this behavior can be used as a basis to triage selective classifications: Documents in the $\hrRegionFull$ region might be treated as automated, or semi-automated, predictions in the decision pipeline, whereas other documents might be triaged by $\sdm(\vz')_{\hat{y}}$ for calling more resource-intensive LM tools, or human adjudication. Importantly, as we discuss below with the covariate-shifted and out-of-distribution datasets, the non-$\sdm$-based estimators provide a less reliable substrate for basing such conditional branching decisions.

\input{tables/table-abbreviated-experiments-classification-main-text.tex}

\paragraph{Covariate-shifted and Out-of-distribution data.} With $\datasetSentimentOOD$, the distinctions among the estimators become clear, with the non-$\sdm$-based estimators performing poorly, even in terms of marginal calibration. That would come as a surprise to end-users, whereas with the $\sdmHR$ estimator, the out-of-distribution documents are more reliably rejected, with the few admitted predictions generally obtaining high conditional accuracies, despite the relatively low accuracies over the test set without selection (see $\estimatorNoReject$ in Table~\ref{tab:experiments-sentiment}). A similar pattern is observed over the $\datasetFactcheck$ dataset in Table~\ref{tab:experiments-main-text-abbreviated}, and Table~\ref{tab:experiments-factcheck} in the Appendix.
\input{tables/table-experiments-classification-over-calibration-set.tex} 
\paragraph{Understanding $\minRescaledSimiliarityForHRRegion$.} For reference, Table~\ref{tab:experiments-classification-over-calibration-set} provides the results over $\calibrationSplit$ for the $\sdm$-based estimators. The value of $\minRescaledSimiliarityForHRRegion$ tends to increase as the accuracy over $\calibrationSplit$ decreases, reflecting a more conservative $\hrRegionFull$ region that admits fewer points. Alg.~\ref{alg:estimate-high-reliability-region} failed to find a finite $\minRescaledSimiliarityForHRRegion$ for $\modelMixtralSDM$ over the $\datasetFactcheck$ calibration set at $\alpha=0.95$, so for reference, we also show the $\hrRegionFull$ region at $\alpha=0.94$, as well as in Tables~\ref{tab:experiments-main-text-abbreviated} and \ref{tab:experiments-factcheck}. In this way, $\minRescaledSimiliarityForHRRegion$ provides a principled, data-driven indicator of the reliability of the estimates, which is interpretable as a simple indicator as to whether the conditional accuracies are, or are not, obtainable over $\calibrationSplit$ at the specified $\alpha$. Similar to conformal estimators, and unlike $\estimatorTempScaling$ and $\estimatorVBLL$ estimators, Alg.~\ref{alg:estimate-high-reliability-region} runs after the parameters of the adaptor layer have been fixed, so this indicator also in effect serves as a check on the optimization process (Eq.~\ref{eq:sdm-loss}) itself.

\paragraph{Understanding $\sdmHR$.} The relative proportion of points in the $\hrRegionFull$ region reflects the data and model, and the uncertainty therein. A high abstention rate at a given $\alpha$ over in-distribution data suggests a need for more (or higher quality) data, a stronger model, and/or a reduction in $\alpha$. For certain applications, it may be informative to construct \textit{nested} $\sdm$ estimators across descending $\alpha$ values by successively excluding any higher $\alpha$ regions when running Alg.~\ref{alg:estimate-high-reliability-region}. At test-time, a point would then be assigned to the region of the most conservative (i.e., closer to 1) $\alpha$ in which it is a member. (See Appendix~\ref{appendix:nested-estimators}.) 

\paragraph{Far out-of-distribution data.} The Appendix (Tables~\ref{tab:experiments-sentiment-shuffled} and \ref{tab:experiments-factcheck-shuffled}) provides results for the shuffled variants. The selection criteria of Eq.~\ref{eq:index-conditional-sdm-estimator} reliably rejects the challenging predictions, whereas the non-$\sdm$-based estimators fare poorly, in general. In this way, the $\sdm$ estimator serves as an effective out-of-distribution detection method. With existing methods, defining an out-of-distribution point has been task-specific, and generally challenging over high-dimensional inputs, typically requiring additional modeling beyond that of the calibration or selection method. In contrast, Eq.~\ref{eq:index-conditional-sdm-estimator} provides a principled approach for determining such cut-offs in a data- and model-driven manner, with minimal hyper-parameters, resulting in a separation of points over which the estimator tends to be reliable (namely, the admitted points) and those over which the estimates themselves tend to be unreliable. 

\section{Conclusion}
We introduced $\sdm$ activation functions and $\sdm$ estimators, which are more robust and interpretable estimators of the predictive uncertainty than those based on the commonly used $\softmax$ function. In this way, $\sdm$ activations provide a principled, data-driven substrate for approaching selective classification, calibration, and out-of-distribution detection with language models.

\section*{Limitations}
The $\sdm$ estimator requires the $\trainSplit$ exemplar vectors at test-time, but the additional compute required for test-time matching is similar to that of commonly used dense retrieval mechanisms with LMs, so the additional overhead is achievable in typical use-cases.

For brevity of presentation, in the main text we omit consideration of the Beta-distributed error term that is a function of the effective sample size of split-conformal coverage \citep{Vovk-2012-ConditionalValidity}. In practice, this term is negligible in the present experiments given $\lvert \calibrationSplit \rvert$ and the resolution of the comparisons. See Appendix~\ref{appendix:effective-sample-size} for a further discussion of analyzing the effective sample size for both the class-conditional and prediction-conditional estimates.

Among points in the $\hrRegionFull$ region, $\sdm$ estimators are relatively robust to covariate shifts that alter the relative proportion of points in the regions partitioned by the \textsc{Similarity}, \textsc{Distance}, and \textsc{Magnitude} signals, as well as label shifts, the latter due to the class-wise thresholds of Alg.~\ref{alg:estimate-high-reliability-region}. For example, in the extreme case, the proportion of points for which $\q=0$ and/or $\dVal=0$ can be arbitrarily larger in the held-out $\testSplit$ compared to that seen in training without altering the calibration of the $\hrRegionFull$ region, ceteris paribus. Of course, it is not possible to maintain calibration over all possible distribution shifts.\footnote{With label shifts, the marginal distribution over labels $P(Y)$ changes, while the conditional distribution $P(X \given Y)$ remains unchanged. With covariate shifts, the marginal distribution over features $P(X)$ changes, while the conditional distribution $P(Y \given X)$ remains unchanged. Label shifts are relatively straightforward to model in the selective classification setting via class-wise CDFs. Absent task-specific information (e.g., samples from the new distribution), controlling for covariate shifts is more complex and requires a partitioning of the high-dimensional space, as with $\sdm$ estimators. ``Concept shifts'', where the conditional distribution $P(Y \given X)$ changes, require information exogenous to the $\sdm$ estimator to model.} However, in contrast, the examined non-$\sdm$-based calibration methods work quite poorly in maintaining calibration in high-probability regions in the presence of even modest distribution shifts, since they marginalize over those partitions, and they also marginalize over the labels. That limits the applicability of such calibration methods in practice. Importantly, due to the dense matching and partitioning of the calibration set, $\sdm$ estimators provide interpretability-by-exemplar: A user can examine the applicable documents in training and those in the applicable partition of the calibration set for further analysis and labeling, as needed.

For the experiments in the present work (\S~\ref{sec:models}), the input $\vh$ to the adaptor layers is a mean-pool over the final-layer hidden states concatenated with the final-layer hidden state of the final sequence position. This achieves the desired behavior observed in the experiments, while enabling relatively efficient training since $\vh$ can be calculated once and cached, with reasonable space requirements, for use across all epochs and $J$ iterations of training. In this case, the notion of convolving over sequence positions is not used, since the kernel/filter-width of the CNN adaptor equals $D$, the dimension of $\vh$. We hypothesize that a learned max-pooling over sequence positions as used in \citet{schmaltz-2021-insights} may be more sample efficient (i.e., higher proportions of points in the $\hrRegionFull$ region, ceteris paribus) than mean-pooling, since the filters learn the aggregation rather than the marginalization over positions with mean-pooling. That would also enable the local-level feature decompositions examined in previous works. However, in that case, naive pre-caching of the frozen hidden states for each token position becomes space prohibitive for long sequences. A practical middle-ground may be to down-sample the position-wise hidden states via summary statistics at the token-level; at semantic boundaries (sentence, paragraph, etc.); and/or at conversation boundaries (prompt, answer, etc.), and to concatenate the resulting vectors with the $\vh$ used in the present experiments. We leave to future higher-resourced experiments to examine the significance of those tradeoffs. 

Contemporary commercial LMs served from cloud APIs may not expose the hidden states of the underlying network. In practice, the output from such models can be cross-encoded with an open-weight model, the composition of which then becomes the model over which to calibrate and by extension, the input to the $\sdm$ estimator. The present work examines the $\sdm$-based calibration behavior with open-weight models to facilitate replication in controlled settings; we leave applied examples with commercial black-box APIs to future work.

\bibliography{sdm}

\appendix
\section{Appendix}
The prompts for the tasks appear in Appendix~\ref{sec:task-prompts}. Expanded versions of the tables in the main text appear in Appendix~\ref{sec:additional-main-text-results}. We provide the results for the far out-of-distribution (OOD) shuffled datasets in Appendix~\ref{sec:far-ood}. Additional implementation details are included in Appendix~\ref{appendix:additional-implementation-details}. Appendix~\ref{appendix:effective-sample-size} provides an approach for analyzing the effective sample size for both the prediction-conditional and class-conditional estimates. 
Appendix~\ref{appendix:updatability} discusses local updatability of the $\sdm$ activation function, highlighting the connection to instance-based metric learners.
Appendix~\ref{appendix:ensemble} proposes an approach for constructing an ensemble over all $J$ models created during training. Appendix~\ref{appendix:vbll-models} provides additional details and comparisons for the experiments with variational Bayesian last-layer estimators \citep{HarrisonEtAl-2024-VBLL}. Appendix~\ref{appendix:aux-experiments} briefly describes additional experiments appearing in the code repository. We close with Appendix~\ref{appendix:nested-estimators}, which provides an approach for constructing nested $\sdm$ estimators. 

\subsection{Prompts}\label{sec:task-prompts}

\paragraph{Sentiment.} For the sentiment datasets, we prompt the LMs for a binary classification (\texttt{Yes} or \texttt{No}) as follows:
\begin{quote}
\texttt{Here is a movie review. <review> \underline{DOCUMENT} </review> Is the sentiment of the movie review positive? Answer Yes if the sentiment is positive. Answer No if the sentiment is negative. Start your response with Yes or No.}
\end{quote}
We replace \texttt{\underline{DOCUMENT}} with the corresponding text for each instance.

\paragraph{Factcheck.} Similarly, for the factcheck datasets, we prompt the LMs for a binary classification (\texttt{Yes} or \texttt{No}) as follows:
\begin{quote}
\texttt{Here is a statement that may contain errors. <statement> \underline{DOCUMENT} </statement> Is the statement true? Answer Yes if the statement is true. Answer No if the statement is false. Start your response with Yes or No.}
\end{quote}
As above, we replace \texttt{\underline{DOCUMENT}} with the corresponding text for each instance.

\subsection{Additional Rows for Table~\ref{tab:experiments-main-text-abbreviated}}\label{sec:additional-main-text-results}

Additional rows for Table~\ref{tab:experiments-main-text-abbreviated} appear in Table~\ref{tab:experiments-sentiment} for the sentiment datasets and Table~\ref{tab:experiments-factcheck} for the factcheck datasets.

\subsection{Far OOD Shuffled Datasets}\label{sec:far-ood}
As discussed in Section~\ref{sec:results} in the main text, Table~\ref{tab:experiments-sentiment-shuffled} shows results for the $\datasetSentimentShuffled$ and $\datasetSentimentOODShuffled$ datasets, and Table~\ref{tab:experiments-factcheck-shuffled} shows results for the $\datasetFactcheckShuffled$ datasets.

In the case of $\datasetSentimentShuffled$ and $\datasetSentimentOODShuffled$, the semantics of the original labels are maintained. This requires the models and estimators to attempt a sentiment classification over the bag-of-words input, or reject the classification. This represents the setting where an LM is given far out-of-distribution input, and additionally provides a control on test-set contamination of the underlying LMs, which due to the shuffling, are relatively unlikely to have seen all of the long, contiguous n-gram sequences from these documents in training or fine-tuning.

In the case of $\datasetFactcheckShuffled$, since the task prompt seeks to classify errors, we set the ground-truth labels of the shuffled counterparts to $y=0$.

\subsection{Additional Implementation Details}\label{appendix:additional-implementation-details}

Replication code is available at the following URL: \url{https://github.com/ReexpressAI/sdm_activations}

We mean center the input to $g$, the 1-D CNN of the $\sdm$ activation layer and the otherwise identical CNN adaptors of the baseline comparison estimators, via the mean and standard deviation over $\trainSplit$.  In all experiments with adaptor layers, $M=1000$ and we use a mini-batch size of 50. We use the Adam optimizer \citep{Kingma-2017-Adam-Optimizer} (without weight decay) with a learning rate of $1 \times 10^{-5}$ for training.

\subsubsection{Implementation of the SDM Activation Function}\label{appendix-sdm-activation-implementation}

The $\sdm$ activation function can be calculated using existing numerically stable $\softmax$ implemenations via the following relation:
\begin{align}\label{eq:sdmActivationRelationToSoftmax}
\sdm(\vz')_i &= \frac{
(2+q)^{\dVal \cdot z'_i}
}{
\sum^C_{c=1}{(2+q)^{\dVal \cdot z'_c}}
} \notag \\
&= \frac{
e^{\log_{e}(2+q) \cdot \dVal \cdot z'_i }
}{
\sum^C_{c=1}{e^{\log_{e}(2+q) \cdot \dVal \cdot z'_c }}
}, 1 \le i \le C
\end{align}
The change of base for the loss can be calculated by multiplying the $\log_{e}$ probabilities from standard optimized $\LogSoftmax$ operations by $\frac{1}{\log_{e}(2+q)}$.

\subsubsection{Implementation of the Empirical CDF Function}\label{appendix:empirical-cdf-implementation}

In the present experiments, the empirical CDF functions are implemented such that the distance quantiles are exclusionary at the boundaries. When $\dNearest=0$, the $1-{\rm{eCDF}}^{\cdot}_{\rm{ca}}( \dNearest ) $ quantile is 1, and when $\dNearest$ is greater than the maximum observed distance (across $\calibrationSplit$ for $\vx \in \testSplit$ and $\vx \in \calibrationSplit$, and across $\trainSplit$ for $\vx \in \trainSplit$, the latter case only occurring during training), the $1-{\rm{eCDF}}^{\cdot}_{\rm{ca}}( \dNearest ) $ quantile is 0.

\subsubsection{Training Dynamics}

As indicated by our loss notation (Eq.~\ref{eq:sdmActivation}) and described in \S~\ref{sec:similarity-calculation} and \S~\ref{sec:distance-calculation}, instance-wise $\q$ and $\dVal$ are not parameters learned directly via gradient descent. By design, we are not back-propagating through a bi- and/or cross-encoded search graph. The KNN approximations of \citet[\S 3.7.1]{schmaltz-2021-insights} are trained with an iterative loss-masking approach to limit overfitting to outliers; with the $\sdm$ activation, instance-wise $\q$ and $\dVal$ serve as the regularizers during learning.

\subsection{Analyzing the Effective Sample Size}\label{appendix:effective-sample-size}

In the context of the $\sdm$ estimator, to parameterize the prior belief that data points with a looser connection to $\trainSplit$ reflect smaller effective sample sizes, while also explicitly accounting for the count of observed points in $\calibrationSplit$, the effective sample size for each test instance can be estimated with the following conservative assumption:

\begin{assumption}\label{assumption-sample-size}
The effective sample size is increasing in $\rescaledSimilarity$, class-wise over $\calibrationSplit$.
\end{assumption}

For each $\vx \in \testSplit$, using $\rescaledSimilarity$, we calculate $\mathbf{\hat{n}}$, the vector of effective sample sizes across classes, relative to $\calibrationSplit$, as: 
\begin{align}\label{equation:effective-sample-size} 
\mathbf{\hat{n}} = &[ \lvert \calibrationSplit \rvert^{y_1} \cdot {\rm{eCDF}}^{y_1}_{\rm{ca}}( \rescaledSimilarity ) , \ldots , \notag \\ &~\lvert \calibrationSplit \rvert^{y_C} \cdot {\rm{eCDF}}^{y_C}_{\rm{ca}}( \rescaledSimilarity ) ]
\end{align}
where $ \lvert \calibrationSplit \rvert^{y_c}$ is the count of calibration set points with true label $y = c$.

The estimate of the effective sample size for each label can then be used to estimate the Beta-distributed error term of split-conformal coverage \citep{Vovk-2012-ConditionalValidity}, providing a sample-size-based error estimate for the class-conditional estimate, assuming exchangeability. 

For the prediction-conditional estimates, assuming independent and identically distributed (i.i.d.) data, these sample size estimates can be used to construct a band around the empirical CDFs over $\dNearest$ (Eq.~\ref{equation:d}) using the sharp constant \citep{Massart-1990-DKW-Tight-Constant} of the distribution-free DKW inequality \citep{DKW-1956-DKW-Inequality}, with $\hat{n}_{\rm{min}}$ taken as the minimum among the estimated sample sizes across classes for the test instance:
\begin{align}\label{equation:dkw} 
\epsilon &= \sqrt{\frac{1}{2 \cdot \hat{n}_{\rm{min}} } \log_e \left ( \frac{2}{1-\alpha} \right )}, \\
\hat{n}_{\rm{min}} &=  \min \left[ \hat{n}_1 , \ldots, \hat{n}_C \right]
\end{align}
If $\hat{n}_{\rm{min}} = 0$, our convention is to set $\epsilon=1$. We then construct the conservative lower and upper counterparts to the distance quantile of Eq.~\ref{equation:d}:
\begin{equation}\label{equation:d-lower} 
\dVal_{\rm{lower}} = \max \left( \dVal - \epsilon, 0 \right)
\end{equation}
\begin{equation}\label{equation:d-upper} 
\dVal_{\rm{upper}} = \min \left( \dVal + \epsilon, 1 \right)
\end{equation}
Eq.~\ref{eq:sdmActivation} can then be calculated by substituting $\dVal_{\rm{lower}}$ and $\dVal_{\rm{upper}}$, in turn, for $\dVal$, resulting in a band around the prediction-conditional estimate, $\sdm(\vz')$:
\begin{align}\label{eq:sdmActivationLower}
\sdm(\vz')^{\rm{lower}}_i &= \notag \\
&\frac{
(2+q)^{\dVal_{\rm{lower}} \cdot z'_i}
}{
\sum^C_{c=1}{(2+q)^{\dVal_{\rm{lower}} \cdot z'_c}}
}, 1 \le i \le C
\end{align}
\begin{align}\label{eq:sdmActivationUpper}
\sdm(\vz')^{\rm{upper}}_i &= \notag \\
&\frac{
(2+q)^{\dVal_{\rm{upper}} \cdot z'_i}
}{
\sum^C_{c=1}{(2+q)^{\dVal_{\rm{upper}} \cdot z'_c}}
}, 1 \le i \le C
\end{align}
A corresponding $\sdmHRLower$ estimator then follows:
\begin{equation}\label{equation:rescaled-similarity-lower} 
\rescaledSimilarityLower = \min \left ( \q, (2+q)^{\sdm(\vz')^{\rm{lower}}_{\hat{y}}} \right )
\end{equation}
\begin{align}\label{eq:index-conditional-sdm-estimator-lower}
&\sdmHRLower := \notag \\
&\begin{cases}
  \hat{y} & \text{if~}  \rescaledSimilarityLower \ge \minRescaledSimiliarityForHRRegion  \wedge \sdm(\vz')^{\rm{lower}}_{\hat{y}} \ge \psi_{\hat{y}}  \\
  \bot & \text{otherwise}
\end{cases}
\end{align}
In the experiments, we focus on the $\sdmHR$ estimator to simplify the presentation, and since a well-calibrated $\sdmHR$ estimator implies a well-calibrated $\sdmHRLower$ estimator. For applications, we recommend taking into account the effective sample size at test-time via $\dVal_{\rm{lower}}$, $\sdm(\vz')^{\rm{lower}}$, $\rescaledSimilarityLower$, and $\sdmHRLower$. The publicly available code (Appendix~\ref{appendix:additional-implementation-details}) calculates these additional values.

\subsection{Updatability}\label{appendix:updatability} 

The $\sdm$ activation function inherits the \textit{updatability} property of instance-based metric learners. Instances with labels $y \in \gY = \{1, \ldots, C\}$ can be dynamically added to $\trainSplit$ after training. This can change the $\Similarity$ and $\Distance$ values, and by extension the uncertainty estimates, while the $\Magnitude$, and by extension the $\argmax$ prediction, remains unchanged provided the weights of the CNN adaptor are held fixed. We hypothesize this can be a useful tradeoff between fast moving weights and slow moving weights in continual learning settings.

Relatedly, instances with labels $y \notin \gY$ can also be added to $\trainSplit$ after training. Given Eq.~\ref{equation:q}, test instances matching to such instances will have reduced $\q$ values, ceteris paribus, since the model can never predict such labels. This is potentially a useful, lightweight alternative to adding an explicit ``out-of-distribution class'' to $\gY$, a comparison to which we leave to future work.

\subsection{Estimator Ensembles}\label{appendix:ensemble} 

At the cost of additional compute, we can control for uncertainty across shuffles of $\trainSplit$ and $\calibrationSplit$ and parameter initializations by constructing an ensemble across all $J$ models saved during training. The publicly available code (Appendix~\ref{appendix:additional-implementation-details}) provides an option for constructing an ensembled $\sdmHRLower$ estimator by only admitting a point if $\hat{y}$ is identical across all $J$ models \textit{and} the point falls into the $\sdmHRLower$ region for all $J$ models. 

\subsection{Comparison to Bayesian Last-layer Networks}\label{appendix:vbll-models}

In this section, we provide additional details for the comparisons to variational Bayesian last-layer neural networks ($\estimatorVBLL$), a computationally efficient Bayesian approach \citep{HarrisonEtAl-2024-VBLL}. The basic setup is similar to the Frequentist and empirically motivated approaches using CNN-based adaptors examined in the main text in that it involves training a small final-layer adaptor over the frozen parameters of the language model. However, in this case, the adaptor network is a multi-layer perceptron (MLP) combined with the $\estimatorVBLL$ estimator. We consider both the discriminative and generative $\estimatorVBLL$ estimators.

\paragraph{Models.} We follow the parameter choices and architecture of \citet{HarrisonEtAl-2024-VBLL}, and its associated code tutorial\footnote{Available at \url{https://github.com/VectorInstitute/vbll}}, with applicable changes to match the experimental settings of the other models and estimators. Specifically, the \textsc{VBLL} models consist of an input linear layer, 2 core linear layers, and a final output linear layer. The hidden dimension is set at 795, which yields a similar number of parameters as the $\sdm$ activation layers (approximately 6 million parameters for the \texttt{Phi-3.5-mini-instruct} adaptors and approximately 8 million parameters for the \texttt{Mixtral 8x7B Instruct v0.1} adaptors). The input to the \textsc{VBLL} models is the same mean-centered embeddings used with the $\sdm$ activation layers and the baseline CNN adaptor layers of the main text. We use the AdamW optimizer \citep{LoshchilovAndHutter-2019-AdamW} with a weight decay of $1 \times 10^{-4}$, which following the code tutorial, is not applied to the final layer. We use a learning rate of $1 \times 10^{-5}$ for training, matching that used with the $\sdm$ activation layers. Following the code tutorial, we use gradient clipping with a max norm of 1, and we use the exponential linear unit (\textsc{ELU}) as the activation function \citep{ClevertEtal-2016-ELU}. We train for 200 epochs, choosing the epoch with the lowest held-out validation loss over $\calibrationSplit$ as the chosen weights. We repeat this process for $J=10$ shuffles of the data (i.e., splits of $\trainSplit$ and $\calibrationSplit$), choosing the model with the globally lowest overall held-out validation loss over $\calibrationSplit$ as the final model.

For the discriminative models over \texttt{Phi-3.5-mini-instruct} and \texttt{Mixtral 8x7B Instruct v0.1}, we use the labels $\modelPhiThreeFiveInstructDiscVBLLMLP$ and $\modelMixtralDiscVBLLMLP$, respectively. Those are the models appearing in Table~\ref{tab:experiments-main-text-abbreviated} in the main text. For the generative models over \texttt{Phi-3.5-mini-instruct} and \texttt{Mixtral 8x7B Instruct v0.1}, we use the labels $\modelPhiThreeFiveInstructGenVBLLMLP$ and $\modelMixtralGenVBLLMLP$, respectively. For all of the aforementioned models, we use a KL regularization weight set at $\frac{1}{|\trainSplit|}$, as in the original paper. We also consider analogous models that increase the KL regularization weight by a multiplicative factor of 50. Increasing the KL regularization weight is suggested in the code tutorial as the ``simplest and most effective method to control the scale of uncertainty'' of $\estimatorVBLL$ models. For these models with the larger KL regularization weight, we use the labels 
$\modelPhiThreeFiveInstructDiscVBLLMLPrFifty$, 
$\modelMixtralDiscVBLLMLPrFifty$, 
$\modelPhiThreeFiveInstructGenVBLLMLPrFifty$, and
$\modelMixtralGenVBLLMLPrFifty$, respectively.

We use the label $\estimatorVBLL$ for the estimator that thresholds the output of the variational Bayesian last-layer neural network at $\alpha$ for the predicted class, analogous to the $\softmax$ and $\sdmAtAlpha$ estimators. The $\estimatorNoReject$ estimator provides a reference point without any selection criteria applied (i.e., the standard marginal and class- and prediction-conditional accuracies over the given dataset).

\paragraph{Results.} The results for the sentiment datasets appear in Table~\ref{tab:experiments-sentiment-additional}, and the results for the factcheck datasets appear in Table~\ref{tab:experiments-factcheck-additional}. For reference, in all cases we also provide the results over the held-out calibration set, $\calibrationSplit$, which is the held-out split used to determine the final model weights, as noted above. 

Comparing to Table~\ref{tab:experiments-classification-over-calibration-set}, the accuracies over $\calibrationSplit$ for the $\estimatorNoReject$ estimators are similar for the $\estimatorVBLL$ models and the models using $\sdm$ activation functions. The MLPs of the $\estimatorVBLL$ models and the 1-D CNNs of the $\sdm$ activation layers have a similar number of parameters and are trained with the same maximum number of epochs and the same number of iterated shuffles of $\trainSplit$ and $\calibrationSplit$. The accuracies without selection for the $\datasetSentiment$ calibration set are already at least $\alpha$, and those for the $\datasetFactcheck$ calibration set are below $\alpha$, but at least $\alpha-0.05$. As such, differences in calibration effectiveness of the respective methods are not directly attributable to substantively different baseline accuracies for the in-distribution tasks.

We find that the $\estimatorVBLL$ estimators are well-calibrated in high-probability regions over in-distribution data, but generally fare poorly over covariate shifts and out-of-distribution data. We find no clear advantages nor disadvantages for the discriminative vs. generative variants, and modifying the KL regularization weight has a minimal impact, at least at this scale. In \citet{HarrisonEtAl-2024-VBLL}, $\estimatorVBLL$ estimators over LMs for sentiment classification are only compared to an MLP baseline on in-distribution data; our evaluation setting is significantly more challenging and closer to real-world conditions encountered with LM applications.

\subsection{Additional Experiments}\label{appendix:aux-experiments}

The publicly available code repository (Appendix~\ref{appendix:additional-implementation-details}) includes additional experiments to further examine the behavior of $\sdm$ activations and estimators. In the interest of length since these experiments are secondary to the main experiments, we defer a full discussion to the code repository.

First, we demonstrate that the behavior of the $\sdm$ estimator is not restricted to binary classification. This is shown by examining the training dynamics, calibration results, and interpretability-by-exemplar behavior on the standard AGNews (4-class classification) dataset from \citet{ZhangEtAl-CharacterCNNforClassification}.

The $\sdm$ estimator is intended to be robust to the optimization process that determines the parameters of the adaptor of the $\sdm$ activation. We examine this with an ablation using the hidden states of the underlying model as the representations used for matching, instead of those of a learned CNN. The result is the $\sdm$ estimator remains well-calibrated at $\alpha$, but with the additional cost of $L^2$ matching with a higher-dimensional $\vh'$, and reduced statistical efficiency, as reflected in fewer points in the $\hrRegionFull$ region, on most of the examined datasets, ceteris paribus. In this way, we demonstrate that Alg.~\ref{alg:estimate-high-reliability-region} is robust to representations held constant during the optimization of Eq.~\ref{eq:sdm-loss}.

\subsection{Nested SDM Estimators}\label{appendix:nested-estimators}

An $\sdm$ estimator calibrated at a given $\alpha$ provides no information in itself of the relative ranking of the probability of the instances outside the $\hrRegionShort$ region. For points falling outside the $\hrRegionShort$ region, one approach is to simply triage by $\sdm(\vz')_{\hat{y}}$ for routing to additional processes in the decision-making pipeline, such as human review. A disadvantage of such relative ranking of the remaining instances is that it does not provide a constraint on the class-conditional accuracy analogous to the selection criteria of Eq.~\ref{eq:index-conditional-sdm-estimator}.

An alternative choice for ranking the instances outside the $\hrRegionShort$ region is to simply run Alg.~\ref{alg:estimate-high-reliability-region} for descending $\alpha$ values, successively excluding the points of each higher region that is found (i.e., with a finite $\minRescaledSimiliarityForHRRegion$). We refer to these as \textit{nested} $\sdm$ estimators to emphasize the use of the residual points. Instances falling in regions partitioned by higher $\alpha$ values are preferable, other things being equal.

Concretely, given a resolution $r \in (0,\frac{1}{2})$, Alg.~\ref{alg:estimate-high-reliability-region} is run with descending values of $\alpha$ spaced by $r$ (starting at $1-r$), successively removing the points in the calibration set, $\calibrationSplit$, for each region recorded with a finite $\minRescaledSimiliarityForHRRegion$. Each region is defined by its $\alpha$, $\minRescaledSimiliarityForHRRegion$, and output thresholds, $[\psi_1, \ldots, \psi_C]$. The user specifies $r$, but not all descending $\alpha$ values may have a corresponding region. The procedure appears in Alg.~\ref{alg:estimate-nested-high-reliability-regions}. 
\input{algorithms/algorithm-estimate-nested-regions.tex}
\paragraph{Test-time.} For new, unseen test instances, the application and behavior of the selection criteria of Eq.~\ref{eq:index-conditional-sdm-estimator} for the most conservative $\hrRegionShort$ region (nearest 1) \textit{is the same as described in the main text}. However, now for points that fall outside that $\hrRegionShort$ region, we perform a search for membership in the highest probability region among the remaining lower probability regions using Eq.~\ref{eq:index-conditional-sdm-estimator}, but with the $\minRescaledSimiliarityForHRRegion$ and $[\psi_1, \ldots, \psi_C]$ values for each remaining region.  

\paragraph{Accounting for the Effective Sample Size.} The application of the DKW inequality on the distance quantiles is the same as in Appendix~\ref{appendix:effective-sample-size}. For the nested estimators, we maintain the convention of pinning the $\alpha$ of Eq.~\ref{equation:dkw} to that of the corresponding region. As such, each point assigned to a region is also assigned to a corresponding (and possibly lower probability) \textit{lower} region using Eq.~\ref{eq:index-conditional-sdm-estimator-lower}, if such a region exists. 

\paragraph{Implementation.} The publicly available code (Appendix~\ref{appendix:additional-implementation-details}) provides a reference implementation. 

\input{tables/table-experiments-classification-sentiment.tex}
\input{tables/table-experiments-classification-factcheck.tex}

\input{tables/table-experiments-classification-sentiment-shuffled.tex}
\input{tables/table-experiments-classification-factcheck-shuffled.tex}

\input{tables/additional/table-experiments-classification-sentiment-additional.tex}
\input{tables/additional/table-experiments-classification-factcheck-additional.tex}

\end{document}

%% file: algorithms/algorithm-estimate-high-reliability-region.tex
\begin{algorithm*}
    \caption{Search Algorithm to Find $\minRescaledSimiliarityForHRRegion$ and $[\psi_1, \ldots, \psi_C]$ to Estimate the $\hrRegionFull$ Region}
    \label{alg:estimate-high-reliability-region}
    \small
    \begin{algorithmic}[1] 
        \Require cached $(\rescaledSimilarity, \sdm(\vz'))$ $\forall ~\vx \in \calibrationSplit$, $\alpha \in (\frac{1}{2},1]$ %
        \Procedure{estimate-high-reliability-region}{cached $(\rescaledSimilarity, \sdm(\vz'))$ $\forall ~\vx \in \calibrationSplit$, $\alpha \in (\frac{1}{2},1]$}
        \State $\minRescaledSimiliarityForHRRegion \gets \infty$ \Comment{A finite $\minRescaledSimiliarityForHRRegion$ may not be found.}
        \State $ [\psi_1, \ldots, \psi_C] \gets [\infty, \ldots, \infty]$ \Comment{Class-wise output thresholds}
        \State $\text{sortedList}~ \gets {\rm{sorted}} ~[\rescaledSimilarity \in \calibrationSplit~{\rm{s.t.}}~\rescaledSimilarity > 0]$ \label{line:ood-hardqbin} \Comment{Restricted to $\rescaledSimilarity > 0$ to exclude OOD}
        \For{$\q^* \in \text{sortedList}$} 
            \State Construct ${\rm{eCDF}}^{y_1}_{\rm{ca}} , \ldots , {\rm{eCDF}}^{y_C}_{\rm{ca}} $ for all $ \rescaledSimilarity \ge \q^*$ in $\calibrationSplit$\label{line:eCDFs} \Comment{eCDFs for $\sdm(\vz')$ (Eq.~\ref{eq:sdmActivation}), stratified by $y$}
            \State Calculate $\psi_c={\rm{inverseCDF}}^{y_c}_{\rm{ca}}(1-\alpha)  ~\forall~c~\in \{1,\ldots,C\}$ \Comment{Quantile functions are inverses of L.~\ref{line:eCDFs}}
	\If{${\rm{all}}( ~ [\psi_1, \ldots, \psi_C] \ge \alpha~)$} \Comment{Element-wise comparison}
        	    \State $\minRescaledSimiliarityForHRRegion \gets \q^*$ 
	    \State \textbf{break}
        \EndIf        
        \EndFor
        \State \textbf{return} $\minRescaledSimiliarityForHRRegion$, $ [\psi_1, \ldots, \psi_C]$
        \EndProcedure
        \Ensure $\minRescaledSimiliarityForHRRegion$, $ [\psi_1, \ldots, \psi_C]$
    \end{algorithmic}
\end{algorithm*}

%% file: tables/table-abbreviated-experiments-classification-main-text.tex
\begin{table*}[h]

  \centering
  \resizebox{0.7\textwidth}{!}{%
\begin{tabular}{l l  l c c c c c c c c c l }
    \toprule

    & & &  \multicolumn{4}{c}{Class-conditional} & \multicolumn{4}{c}{Prediction-conditional}  & \multicolumn{2}{c}{Marginal} \\
    & & &  \multicolumn{2}{c}{$y=0$} & \multicolumn{2}{c}{$y=1$} & \multicolumn{2}{c}{$\hat{y}=0$} & \multicolumn{2}{c}{$\hat{y}=1$} & \multicolumn{2}{c}{$y\in \{0,1\}$} \\
    \cmidrule(r){4-5} \cmidrule(r){6-7} \cmidrule(r){8-9} \cmidrule(r){10-11} \cmidrule(r){12-13} \\
    Dataset   & Model & Estimator & \textsc{Acc.}& $\frac{n}{|\testSplit|}$ & \textsc{Acc.} & $\frac{n}{|\testSplit|}$ & \textsc{Acc.} & $\frac{n}{|\testSplit|}$ & \textsc{Acc.} & $\frac{n}{|\testSplit|}$ & \textsc{Acc.} & $\frac{n}{|\testSplit|}$\\
  \midrule 
 %
$\datasetSentiment$ & $\modelPhiThreeFiveInstruct$ & $\estimatorSoftmax$ & \colorbox{correctPredictionColor}{0.98} & 0.50 & \colorbox{wrongPredictionColor}{0.86} & 0.48 & \colorbox{wrongPredictionColor}{0.88} & 0.56 & \colorbox{correctPredictionColor}{0.98} & 0.42 & \colorbox{wrongPredictionColor}{0.93} & 0.98\\
$\datasetSentiment$ & $\modelPhiThreeFiveInstruct$ & $\estimatorTempScaling$ & \colorbox{correctPredictionColor}{0.99} & 0.49 & \colorbox{wrongPredictionColor}{0.91} & 0.41 & \colorbox{wrongPredictionColor}{0.93} & 0.52 & \colorbox{correctPredictionColor}{0.99} & 0.38 & \colorbox{correctPredictionColor}{0.95} & 0.90\\
$\datasetSentiment$ & $\modelPhiThreeFiveInstruct$ & $\conformalAPS$ & \colorbox{correctPredictionColor}{0.99} & 0.49 & \colorbox{wrongPredictionColor}{0.92} & 0.40 & \colorbox{wrongPredictionColor}{0.94} & 0.51 & \colorbox{correctPredictionColor}{0.99} & 0.37 & \colorbox{correctPredictionColor}{0.96} & 0.89\\
\bottomrule
  \end{tabular}
  }  
    \caption{Comparison of estimators for the sentiment and factcheck datasets, with \colorbox{correctPredictionColor}{$\alpha$}$=0.95$. \colorbox{correctPredictionColor}{\allRejected} indicates all predictions were rejected, which is preferred over falling \colorbox{wrongPredictionColor}{under} the expected accuracy. 
  $n=|\text{Admitted}|$, the count of non-rejected documents. The rows corresponding to the proposed $\sdm$ estimator (Eq.~\ref{eq:index-conditional-sdm-estimator}) are \colorbox{lightestgray}{highlighted}.} %
  \label{tab:experiments-main-text-abbreviated} 
\end{table*}

%% file: tables/table-experiments-classification-over-calibration-set.tex
\begin{table*}

  \centering
  \resizebox{1.0\textwidth}{!}{%
  \begin{tabular}{l l  l c c c c c c c c c l }
    \toprule

    & & &  \multicolumn{4}{c}{Class-conditional} & \multicolumn{4}{c}{Prediction-conditional}  & \multicolumn{2}{c}{Marginal} \\
    & & &  \multicolumn{2}{c}{$y=0$} & \multicolumn{2}{c}{$y=1$} & \multicolumn{2}{c}{$\hat{y}=0$} & \multicolumn{2}{c}{$\hat{y}=1$} & \multicolumn{2}{c}{$y\in \{0,1\}$} \\
    \cmidrule(r){4-5} \cmidrule(r){6-7} \cmidrule(r){8-9} \cmidrule(r){10-11} \cmidrule(r){12-13} \\
    Dataset   & Model & Estimator & \textsc{Acc.}& $\frac{n}{|\calibrationSplit|}$ & \textsc{Acc.} & $\frac{n}{|\calibrationSplit|}$ & \textsc{Acc.} & $\frac{n}{|\calibrationSplit|}$ & \textsc{Acc.} & $\frac{n}{|\calibrationSplit|}$ & \textsc{Acc.} & $\frac{n}{|\calibrationSplit|}$\\
  \midrule 
 $\datasetSentiment \calibrationSplit$ & $\modelPhiThreeFiveInstructSDM$ & $\estimatorNoReject$ & \colorbox{correctPredictionColor}{0.95} & 0.50 & \colorbox{correctPredictionColor}{0.96} & 0.50 & \colorbox{correctPredictionColor}{0.96} & 0.50 & \colorbox{correctPredictionColor}{0.95} & 0.50 & \colorbox{correctPredictionColor}{0.96} & 1.\\
$\datasetSentiment \calibrationSplit$ & $\modelPhiThreeFiveInstructSDM$ & $\estimatorSoftmax$ & \colorbox{correctPredictionColor}{0.96} & 0.48 & \colorbox{correctPredictionColor}{0.97} & 0.48 & \colorbox{correctPredictionColor}{0.97} & 0.47 & \colorbox{correctPredictionColor}{0.96} & 0.49 & \colorbox{correctPredictionColor}{0.97} & 0.96\\
$\datasetSentiment \calibrationSplit$ & $\modelPhiThreeFiveInstructSDM$ & $\estimatorSoftmaxOverDistanceMagnitude$ & \colorbox{correctPredictionColor}{0.99} & 0.31 & \colorbox{correctPredictionColor}{0.99} & 0.24 & \colorbox{correctPredictionColor}{1.00} & 0.31 & \colorbox{correctPredictionColor}{0.99} & 0.24 & \colorbox{correctPredictionColor}{0.99} & 0.55\\
$\datasetSentiment \calibrationSplit$ & $\modelPhiThreeFiveInstructSDM$ & $\sdmAtAlpha$ & \colorbox{correctPredictionColor}{0.99} & 0.42 & \colorbox{correctPredictionColor}{0.99} & 0.39 & \colorbox{correctPredictionColor}{0.99} & 0.42 & \colorbox{correctPredictionColor}{0.99} & 0.39 & \colorbox{correctPredictionColor}{0.99} & 0.81\\
$\datasetSentiment \calibrationSplit$ & $\modelPhiThreeFiveInstructSDM$ & $\sdmHR, \alpha=0.95, \textcolor{blueEmphasis}{\minRescaledSimiliarityForHRRegion=52.2}$ & \colorbox{correctPredictionColor}{0.99} & 0.38 & \colorbox{correctPredictionColor}{0.99} & 0.32 & \colorbox{correctPredictionColor}{0.99} & 0.38 & \colorbox{correctPredictionColor}{0.99} & 0.31 & \colorbox{correctPredictionColor}{0.99} & 0.69\\
\midrule
$\datasetSentiment \calibrationSplit$ & $\modelMixtralSDM$ & $\estimatorNoReject$ & \colorbox{correctPredictionColor}{0.96} & 0.50 & \colorbox{correctPredictionColor}{0.96} & 0.50 & \colorbox{correctPredictionColor}{0.96} & 0.50 & \colorbox{correctPredictionColor}{0.96} & 0.50 & \colorbox{correctPredictionColor}{0.96} & 1.\\
$\datasetSentiment \calibrationSplit$ & $\modelMixtralSDM$ & $\estimatorSoftmax$ & \colorbox{correctPredictionColor}{0.96} & 0.49 & \colorbox{correctPredictionColor}{0.97} & 0.49 & \colorbox{correctPredictionColor}{0.97} & 0.49 & \colorbox{correctPredictionColor}{0.96} & 0.49 & \colorbox{correctPredictionColor}{0.97} & 0.98\\
$\datasetSentiment \calibrationSplit$ & $\modelMixtralSDM$ & $\estimatorSoftmaxOverDistanceMagnitude$ & \colorbox{correctPredictionColor}{1.00} & 0.43 & \colorbox{correctPredictionColor}{0.99} & 0.35 & \colorbox{correctPredictionColor}{0.99} & 0.43 & \colorbox{correctPredictionColor}{1.00} & 0.34 & \colorbox{correctPredictionColor}{0.99} & 0.78\\
$\datasetSentiment \calibrationSplit$ & $\modelMixtralSDM$ & $\sdmAtAlpha$ & \colorbox{correctPredictionColor}{0.99} & 0.47 & \colorbox{correctPredictionColor}{0.98} & 0.43 & \colorbox{correctPredictionColor}{0.98} & 0.47 & \colorbox{correctPredictionColor}{0.98} & 0.43 & \colorbox{correctPredictionColor}{0.98} & 0.90\\
$\datasetSentiment \calibrationSplit$ & $\modelMixtralSDM$ & $\sdmHR, \alpha=0.95, \textcolor{blueEmphasis}{\minRescaledSimiliarityForHRRegion=63.0}$ & \colorbox{correctPredictionColor}{0.99} & 0.41 & \colorbox{correctPredictionColor}{0.99} & 0.34 & \colorbox{correctPredictionColor}{0.99} & 0.41 & \colorbox{correctPredictionColor}{0.99} & 0.34 & \colorbox{correctPredictionColor}{0.99} & 0.74\\

  \midrule 
$\datasetFactcheck \calibrationSplit$ & $\modelPhiThreeFiveInstructSDM$ & $\estimatorNoReject$ & \colorbox{wrongPredictionColor}{0.90} & 0.50 & \colorbox{wrongPredictionColor}{0.91} & 0.50 & \colorbox{wrongPredictionColor}{0.91} & 0.49 & \colorbox{wrongPredictionColor}{0.90} & 0.51 & \colorbox{wrongPredictionColor}{0.90} & 1.\\
$\datasetFactcheck \calibrationSplit$ & $\modelPhiThreeFiveInstructSDM$ & $\estimatorSoftmax$ & \colorbox{wrongPredictionColor}{0.94} & 0.32 & \colorbox{correctPredictionColor}{0.96} & 0.41 & \colorbox{correctPredictionColor}{0.95} & 0.31 & \colorbox{correctPredictionColor}{0.96} & 0.41 & \colorbox{correctPredictionColor}{0.96} & 0.72\\
$\datasetFactcheck \calibrationSplit$ & $\modelPhiThreeFiveInstructSDM$ & $\estimatorSoftmaxOverDistanceMagnitude$ & \colorbox{correctPredictionColor}{0.98} & 0.08 & \colorbox{correctPredictionColor}{1.00} & 0.07 & \colorbox{correctPredictionColor}{1.00} & 0.08 & \colorbox{correctPredictionColor}{0.98} & 0.07 & \colorbox{correctPredictionColor}{0.99} & 0.15\\
$\datasetFactcheck \calibrationSplit$ & $\modelPhiThreeFiveInstructSDM$ & $\sdmAtAlpha$ & \colorbox{correctPredictionColor}{0.98} & 0.33 & \colorbox{correctPredictionColor}{0.99} & 0.27 & \colorbox{correctPredictionColor}{0.99} & 0.33 & \colorbox{correctPredictionColor}{0.98} & 0.27 & \colorbox{correctPredictionColor}{0.98} & 0.60\\
$\datasetFactcheck \calibrationSplit$ & $\modelPhiThreeFiveInstructSDM$ & $\sdmHR, \alpha=0.95, \textcolor{blueEmphasis}{\minRescaledSimiliarityForHRRegion=95.0}$ & \colorbox{correctPredictionColor}{1.00} & 0.19 & \colorbox{correctPredictionColor}{0.99} & 0.12 & \colorbox{correctPredictionColor}{1.00} & 0.19 & \colorbox{correctPredictionColor}{0.99} & 0.12 & \colorbox{correctPredictionColor}{1.00} & 0.31\\
\midrule
$\datasetFactcheck \calibrationSplit$ & $\modelMixtralSDM$ & $\estimatorNoReject$ & \colorbox{wrongPredictionColor}{0.90} & 0.50 & \colorbox{wrongPredictionColor}{0.92} & 0.50 & \colorbox{wrongPredictionColor}{0.92} & 0.49 & \colorbox{wrongPredictionColor}{0.90} & 0.51 & \colorbox{wrongPredictionColor}{0.91} & 1.\\
$\datasetFactcheck \calibrationSplit$ & $\modelMixtralSDM$ & $\estimatorSoftmax$ & \colorbox{wrongPredictionColor}{0.94} & 0.44 & \colorbox{correctPredictionColor}{0.96} & 0.45 & \colorbox{correctPredictionColor}{0.96} & 0.43 & \colorbox{wrongPredictionColor}{0.94} & 0.46 & \colorbox{correctPredictionColor}{0.95} & 0.89\\
$\datasetFactcheck \calibrationSplit$ & $\modelMixtralSDM$ & $\estimatorSoftmaxOverDistanceMagnitude$ & \colorbox{correctPredictionColor}{0.98} & 0.28 & \colorbox{correctPredictionColor}{0.98} & 0.17 & \colorbox{correctPredictionColor}{0.99} & 0.27 & \colorbox{correctPredictionColor}{0.96} & 0.17 & \colorbox{correctPredictionColor}{0.98} & 0.44\\
$\datasetFactcheck \calibrationSplit$ & $\modelMixtralSDM$ & $\sdmAtAlpha$ & \colorbox{correctPredictionColor}{0.97} & 0.41 & \colorbox{correctPredictionColor}{0.95} & 0.32 & \colorbox{correctPredictionColor}{0.96} & 0.41 & \colorbox{correctPredictionColor}{0.95} & 0.32 & \colorbox{correctPredictionColor}{0.96} & 0.73\\
$\datasetFactcheck \calibrationSplit$ & $\modelMixtralSDM$ & $\sdmHR, \alpha=0.95, \textcolor{blueEmphasis}{\minRescaledSimiliarityForHRRegion=\infty}$ & \colorbox{correctPredictionColor}{\allRejected} & 0. & \colorbox{correctPredictionColor}{\allRejected} & 0. & \colorbox{correctPredictionColor}{\allRejected} & 0. & \colorbox{correctPredictionColor}{\allRejected} & 0. & \colorbox{correctPredictionColor}{\allRejected} & 0.\\
$\datasetFactcheck \calibrationSplit$ & $\modelMixtralSDM$ & $\sdmAtAlpha, \alpha=0.94$ & \colorbox{correctPredictionColor}{0.96} & 0.42 & \colorbox{correctPredictionColor}{0.95} & 0.33 & \colorbox{correctPredictionColor}{0.96} & 0.42 & \colorbox{correctPredictionColor}{0.95} & 0.32 & \colorbox{correctPredictionColor}{0.96} & 0.74\\
$\datasetFactcheck \calibrationSplit$ & $\modelMixtralSDM$ & $\sdmHR, \alpha=0.94, \textcolor{blueEmphasis}{\minRescaledSimiliarityForHRRegion=134.0}$ & \colorbox{correctPredictionColor}{1.00} & 0.33 & \colorbox{correctPredictionColor}{0.96} & 0.14 & \colorbox{correctPredictionColor}{0.99} & 0.33 & \colorbox{correctPredictionColor}{0.99} & 0.13 & \colorbox{correctPredictionColor}{0.99} & 0.46\\

    \bottomrule
  \end{tabular}
  }  
    \caption{Reference results over $\calibrationSplit$ to illustrate the behavior of \textcolor{blueEmphasis}{$\minRescaledSimiliarityForHRRegion$}. The value of $\minRescaledSimiliarityForHRRegion$ tends to increase as the accuracy over $\calibrationSplit$ decreases, reflecting a more conservative $\hrRegionFull$ region. Alg.~\ref{alg:estimate-high-reliability-region} failed to find a finite $\minRescaledSimiliarityForHRRegion$ for $\modelMixtralSDM$ over the $\datasetFactcheck$ calibration set at $\alpha=0.95$, so for reference, we also show the $\hrRegionFull$ region at $\alpha=0.94$. } %
  \label{tab:experiments-classification-over-calibration-set}
\end{table*}

%% file: algorithms/algorithm-estimate-nested-regions.tex
\begin{algorithm*}
    \caption{Descending $\alpha$ Search Algorithm to Estimate Nested Regions}
    \label{alg:estimate-nested-high-reliability-regions}
    \small
    \begin{algorithmic}[1]
        \Require cached $(\rescaledSimilarity, \sdm(\vz'))$ $\forall ~\vx \in \calibrationSplit$, resolution $r \in (0,\frac{1}{2})$ 
        \Procedure{estimate-nested-regions}{cached $(\rescaledSimilarity, \sdm(\vz'))$ $\forall ~\vx \in \calibrationSplit$, $r \in (0,\frac{1}{2})$}
        \State $\gR \gets [~]$ \Comment{Ordered list of the $(\alpha, \minRescaledSimiliarityForHRRegion, [\psi_1, \ldots, \psi_C])$ triples of the recorded regions}
        \State $l \gets 1$
        \State $\calibrationSplit^{(l)} \gets \calibrationSplit$
        \While{$1 - r \cdot l > \frac{1}{2}$} \Comment{$\alpha_l \in (\frac{1}{2},1]$, as required by Alg.~\ref{alg:estimate-high-reliability-region}}
            \State $\alpha_l \gets 1 - r \cdot l$
            \State ${\minRescaledSimiliarityForHRRegion}^{(l)}, [\psi^{(l)}_1, \ldots, \psi^{(l)}_C] \gets$ \Call{estimate-high-reliability-region}{cached $(\rescaledSimilarity, \sdm(\vz'))$ $\forall ~\vx \in \calibrationSplit^{(l)}$, $\alpha_l$} \Comment{Alg.~\ref{alg:estimate-high-reliability-region}}
            \If{${\minRescaledSimiliarityForHRRegion}^{(l)} < \infty$}
                \State Append $(\alpha_l, {\minRescaledSimiliarityForHRRegion}^{(l)}, [\psi^{(l)}_1, \ldots, \psi^{(l)}_C])$ to $\gR$
                \State $\gT^{(l)} \gets \{ \vx \in \calibrationSplit^{(l)} ~{\rm{s.t.}}~ \rescaledSimilarity \ge {\minRescaledSimiliarityForHRRegion}^{(l)} \wedge \{c : \sdm(\vz')_c \ge \psi^{(l)}_c\} = \{\hat{y}\} \}$ \label{line:nested-region-members} \Comment{Region membership is determined via the selection criteria of Eq.~\ref{eq:index-conditional-sdm-estimator}, with $\hat{y} = \argmax_c \vz'_c$.}
                \State $\calibrationSplit^{(l+1)} \gets \calibrationSplit^{(l)} ~\setminus~ \gT^{(l)}$ \Comment{Remove the admitted instances before the next lower $\alpha$.}
            \Else
                \State $\calibrationSplit^{(l+1)} \gets \calibrationSplit^{(l)}$ \Comment{No instances are excluded at an $\alpha_l$ without a finite ${\minRescaledSimiliarityForHRRegion}^{(l)}$.}
            \EndIf
            \State $l \gets l + 1$
        \EndWhile
        \State \textbf{return} $\gR$
        \EndProcedure
        \Ensure $\gR$, which contains the $(\alpha, \minRescaledSimiliarityForHRRegion, [\psi_1, \ldots, \psi_C])$ triples of the recorded regions (descending in $\alpha$)
    \end{algorithmic}
\end{algorithm*}

%% file: tables/table-experiments-classification-sentiment.tex
\begin{table*}

  \centering
  \resizebox{1.0\textwidth}{!}{%
  \begin{tabular}{l l  l c c c c c c c c c l }
    \toprule

    & & &  \multicolumn{4}{c}{Class-conditional} & \multicolumn{4}{c}{Prediction-conditional}  & \multicolumn{2}{c}{Marginal} \\
    & & &  \multicolumn{2}{c}{$y=0$} & \multicolumn{2}{c}{$y=1$} & \multicolumn{2}{c}{$\hat{y}=0$} & \multicolumn{2}{c}{$\hat{y}=1$} & \multicolumn{2}{c}{$y\in \{0,1\}$} \\
    \cmidrule(r){4-5} \cmidrule(r){6-7} \cmidrule(r){8-9} \cmidrule(r){10-11} \cmidrule(r){12-13} \\
    Dataset   & Model & Estimator & \textsc{Acc.}& $\frac{n}{|\testSplit|}$ & \textsc{Acc.} & $\frac{n}{|\testSplit|}$ & \textsc{Acc.} & $\frac{n}{|\testSplit|}$ & \textsc{Acc.} & $\frac{n}{|\testSplit|}$ & \textsc{Acc.} & $\frac{n}{|\testSplit|}$\\
  \midrule 
$\datasetSentiment$ & $\modelPhiThreeFiveInstruct$ & $\estimatorNoReject$ & \colorbox{correctPredictionColor}{0.98} & 0.50 & \colorbox{wrongPredictionColor}{0.85} & 0.50 & \colorbox{wrongPredictionColor}{0.86} & 0.57 & \colorbox{correctPredictionColor}{0.98} & 0.43 & \colorbox{wrongPredictionColor}{0.91} & 1.\\
$\datasetSentiment$ & $\modelPhiThreeFiveInstruct$ & $\estimatorSoftmax$ & \colorbox{correctPredictionColor}{0.98} & 0.50 & \colorbox{wrongPredictionColor}{0.86} & 0.48 & \colorbox{wrongPredictionColor}{0.88} & 0.56 & \colorbox{correctPredictionColor}{0.98} & 0.42 & \colorbox{wrongPredictionColor}{0.93} & 0.98\\
$\datasetSentiment$ & $\modelPhiThreeFiveInstruct$ & $\estimatorTempScaling$ & \colorbox{correctPredictionColor}{0.99} & 0.49 & \colorbox{wrongPredictionColor}{0.91} & 0.41 & \colorbox{wrongPredictionColor}{0.93} & 0.52 & \colorbox{correctPredictionColor}{0.99} & 0.38 & \colorbox{correctPredictionColor}{0.95} & 0.90\\
$\datasetSentiment$ & $\modelPhiThreeFiveInstruct$ & $\conformalAPS$ & \colorbox{correctPredictionColor}{0.99} & 0.49 & \colorbox{wrongPredictionColor}{0.92} & 0.40 & \colorbox{wrongPredictionColor}{0.94} & 0.51 & \colorbox{correctPredictionColor}{0.99} & 0.37 & \colorbox{correctPredictionColor}{0.96} & 0.89\\
 
$\datasetSentiment$ & $\modelPhiThreeFiveInstruct$ & $\conformalRAPS$ & \colorbox{correctPredictionColor}{0.99} & 0.48 & \colorbox{wrongPredictionColor}{0.91} & 0.41 & \colorbox{wrongPredictionColor}{0.93} & 0.51 & \colorbox{correctPredictionColor}{0.99} & 0.38 & \colorbox{correctPredictionColor}{0.95} & 0.90\\
 
$\datasetSentiment$ & $\modelPhiThreeFiveInstructCNNAdaptor$ & $\estimatorNoReject$ & \colorbox{correctPredictionColor}{0.97} & 0.50 & \colorbox{correctPredictionColor}{0.95} & 0.50 & \colorbox{correctPredictionColor}{0.96} & 0.51 & \colorbox{correctPredictionColor}{0.97} & 0.49 & \colorbox{correctPredictionColor}{0.96} & 1.\\
$\datasetSentiment$ & $\modelPhiThreeFiveInstructCNNAdaptor$ & $\estimatorSoftmax$ & \colorbox{correctPredictionColor}{0.99} & 0.42 & \colorbox{correctPredictionColor}{1.00} & 0.42 & \colorbox{correctPredictionColor}{1.00} & 0.42 & \colorbox{correctPredictionColor}{0.99} & 0.42 & \colorbox{correctPredictionColor}{0.99} & 0.84\\
$\datasetSentiment$ & $\modelPhiThreeFiveInstructCNNAdaptor$ & $\estimatorTempScaling$ & \colorbox{correctPredictionColor}{0.99} & 0.42 & \colorbox{correctPredictionColor}{1.00} & 0.41 & \colorbox{correctPredictionColor}{1.00} & 0.42 & \colorbox{correctPredictionColor}{0.99} & 0.41 & \colorbox{correctPredictionColor}{0.99} & 0.83\\
$\datasetSentiment$ & $\modelPhiThreeFiveInstructCNNAdaptor$ & $\conformalAPS$ & \colorbox{correctPredictionColor}{0.98} & 0.45 & \colorbox{correctPredictionColor}{0.98} & 0.45 & \colorbox{correctPredictionColor}{0.98} & 0.45 & \colorbox{correctPredictionColor}{0.98} & 0.45 & \colorbox{correctPredictionColor}{0.98} & 0.90\\
 
$\datasetSentiment$ & $\modelPhiThreeFiveInstructCNNAdaptor$ & $\conformalRAPS$ & \colorbox{correctPredictionColor}{0.98} & 0.45 & \colorbox{correctPredictionColor}{0.98} & 0.44 & \colorbox{correctPredictionColor}{0.98} & 0.45 & \colorbox{correctPredictionColor}{0.98} & 0.44 & \colorbox{correctPredictionColor}{0.98} & 0.89\\
 
$\datasetSentiment$ & $\modelPhiThreeFiveInstructSDM$ & $\estimatorNoReject$ & \colorbox{correctPredictionColor}{0.96} & 0.50 & \colorbox{correctPredictionColor}{0.96} & 0.50 & \colorbox{correctPredictionColor}{0.96} & 0.50 & \colorbox{correctPredictionColor}{0.96} & 0.50 & \colorbox{correctPredictionColor}{0.96} & 1.\\
$\datasetSentiment$ & $\modelPhiThreeFiveInstructSDM$ & $\estimatorSoftmax$ & \colorbox{correctPredictionColor}{0.97} & 0.48 & \colorbox{correctPredictionColor}{0.97} & 0.48 & \colorbox{correctPredictionColor}{0.97} & 0.48 & \colorbox{correctPredictionColor}{0.97} & 0.48 & \colorbox{correctPredictionColor}{0.97} & 0.96\\
$\datasetSentiment$ & $\modelPhiThreeFiveInstructSDM$ & $\estimatorSoftmaxOverDistanceMagnitude$ & \colorbox{correctPredictionColor}{0.99} & 0.30 & \colorbox{correctPredictionColor}{0.99} & 0.24 & \colorbox{correctPredictionColor}{1.00} & 0.30 & \colorbox{correctPredictionColor}{0.99} & 0.24 & \colorbox{correctPredictionColor}{0.99} & 0.54\\
$\datasetSentiment$ & $\modelPhiThreeFiveInstructSDM$ & $\sdmAtAlpha$ & \colorbox{correctPredictionColor}{0.99} & 0.43 & \colorbox{correctPredictionColor}{0.99} & 0.38 & \colorbox{correctPredictionColor}{0.99} & 0.43 & \colorbox{correctPredictionColor}{0.99} & 0.38 & \colorbox{correctPredictionColor}{0.99} & 0.81\\
$\datasetSentiment$ & $\modelPhiThreeFiveInstructSDM$ & $\sdmHR$ & \colorbox{correctPredictionColor}{1.00} & 0.37 & \colorbox{correctPredictionColor}{0.99} & 0.30 & \colorbox{correctPredictionColor}{0.99} & 0.38 & \colorbox{correctPredictionColor}{1.00} & 0.30 & \colorbox{correctPredictionColor}{0.99} & 0.68\\

  \midrule 
$\datasetSentiment$ & $\modelMixtral$ & $\estimatorNoReject$ & \colorbox{correctPredictionColor}{0.98} & 0.50 & \colorbox{wrongPredictionColor}{0.88} & 0.50 & \colorbox{wrongPredictionColor}{0.89} & 0.55 & \colorbox{correctPredictionColor}{0.98} & 0.45 & \colorbox{wrongPredictionColor}{0.93} & 1.\\
$\datasetSentiment$ & $\modelMixtral$ & $\estimatorSoftmax$ & \colorbox{correctPredictionColor}{0.98} & 0.50 & \colorbox{wrongPredictionColor}{0.88} & 0.50 & \colorbox{wrongPredictionColor}{0.89} & 0.55 & \colorbox{correctPredictionColor}{0.98} & 0.45 & \colorbox{wrongPredictionColor}{0.93} & 1.00\\
$\datasetSentiment$ & $\modelMixtral$ & $\estimatorTempScaling$ & \colorbox{correctPredictionColor}{0.99} & 0.50 & \colorbox{wrongPredictionColor}{0.90} & 0.48 & \colorbox{wrongPredictionColor}{0.91} & 0.54 & \colorbox{correctPredictionColor}{0.98} & 0.44 & \colorbox{wrongPredictionColor}{0.94} & 0.98\\
$\datasetSentiment$ & $\modelMixtral$ & $\conformalAPS$ & \colorbox{correctPredictionColor}{0.98} & 0.49 & \colorbox{wrongPredictionColor}{0.91} & 0.47 & \colorbox{wrongPredictionColor}{0.92} & 0.52 & \colorbox{correctPredictionColor}{0.98} & 0.44 & \colorbox{correctPredictionColor}{0.95} & 0.96\\
 
$\datasetSentiment$ & $\modelMixtral$ & $\conformalRAPS$ & \colorbox{correctPredictionColor}{0.99} & 0.49 & \colorbox{wrongPredictionColor}{0.92} & 0.47 & \colorbox{wrongPredictionColor}{0.93} & 0.52 & \colorbox{correctPredictionColor}{0.98} & 0.44 & \colorbox{correctPredictionColor}{0.95} & 0.96\\
 
$\datasetSentiment$ & $\modelMixtralCNNAdaptor$ & $\estimatorNoReject$ & \colorbox{correctPredictionColor}{0.97} & 0.50 & \colorbox{correctPredictionColor}{0.96} & 0.50 & \colorbox{correctPredictionColor}{0.96} & 0.51 & \colorbox{correctPredictionColor}{0.97} & 0.49 & \colorbox{correctPredictionColor}{0.97} & 1.\\
$\datasetSentiment$ & $\modelMixtralCNNAdaptor$ & $\estimatorSoftmax$ & \colorbox{correctPredictionColor}{0.99} & 0.45 & \colorbox{correctPredictionColor}{0.99} & 0.43 & \colorbox{correctPredictionColor}{0.99} & 0.45 & \colorbox{correctPredictionColor}{0.99} & 0.43 & \colorbox{correctPredictionColor}{0.99} & 0.87\\
$\datasetSentiment$ & $\modelMixtralCNNAdaptor$ & $\estimatorTempScaling$ & \colorbox{correctPredictionColor}{0.99} & 0.43 & \colorbox{correctPredictionColor}{0.99} & 0.41 & \colorbox{correctPredictionColor}{0.99} & 0.43 & \colorbox{correctPredictionColor}{0.99} & 0.41 & \colorbox{correctPredictionColor}{0.99} & 0.84\\
$\datasetSentiment$ & $\modelMixtralCNNAdaptor$ & $\conformalAPS$ & \colorbox{correctPredictionColor}{0.99} & 0.46 & \colorbox{correctPredictionColor}{0.98} & 0.45 & \colorbox{correctPredictionColor}{0.98} & 0.46 & \colorbox{correctPredictionColor}{0.99} & 0.44 & \colorbox{correctPredictionColor}{0.99} & 0.91\\
 
$\datasetSentiment$ & $\modelMixtralCNNAdaptor$ & $\conformalRAPS$ & \colorbox{correctPredictionColor}{0.99} & 0.46 & \colorbox{correctPredictionColor}{0.98} & 0.45 & \colorbox{correctPredictionColor}{0.98} & 0.47 & \colorbox{correctPredictionColor}{0.98} & 0.45 & \colorbox{correctPredictionColor}{0.98} & 0.92\\
 
$\datasetSentiment$ & $\modelMixtralSDM$ & $\estimatorNoReject$ & \colorbox{correctPredictionColor}{0.96} & 0.50 & \colorbox{correctPredictionColor}{0.95} & 0.50 & \colorbox{correctPredictionColor}{0.95} & 0.51 & \colorbox{correctPredictionColor}{0.96} & 0.49 & \colorbox{correctPredictionColor}{0.96} & 1.\\
$\datasetSentiment$ & $\modelMixtralSDM$ & $\estimatorSoftmax$ & \colorbox{correctPredictionColor}{0.97} & 0.49 & \colorbox{correctPredictionColor}{0.96} & 0.49 & \colorbox{correctPredictionColor}{0.96} & 0.50 & \colorbox{correctPredictionColor}{0.97} & 0.49 & \colorbox{correctPredictionColor}{0.97} & 0.98\\
$\datasetSentiment$ & $\modelMixtralSDM$ & $\estimatorSoftmaxOverDistanceMagnitude$ & \colorbox{correctPredictionColor}{0.99} & 0.43 & \colorbox{correctPredictionColor}{0.99} & 0.33 & \colorbox{correctPredictionColor}{0.99} & 0.43 & \colorbox{correctPredictionColor}{0.99} & 0.33 & \colorbox{correctPredictionColor}{0.99} & 0.77\\
$\datasetSentiment$ & $\modelMixtralSDM$ & $\sdmAtAlpha$ & \colorbox{correctPredictionColor}{0.98} & 0.48 & \colorbox{correctPredictionColor}{0.98} & 0.43 & \colorbox{correctPredictionColor}{0.98} & 0.47 & \colorbox{correctPredictionColor}{0.98} & 0.43 & \colorbox{correctPredictionColor}{0.98} & 0.90\\
$\datasetSentiment$ & $\modelMixtralSDM$ & $\sdmHR$ & \colorbox{correctPredictionColor}{0.99} & 0.41 & \colorbox{correctPredictionColor}{0.98} & 0.33 & \colorbox{correctPredictionColor}{0.99} & 0.41 & \colorbox{correctPredictionColor}{0.98} & 0.33 & \colorbox{correctPredictionColor}{0.99} & 0.74\\

\midrule
$\datasetSentimentOOD$ & $\modelPhiThreeFiveInstruct$ & $\estimatorNoReject$ & \colorbox{correctPredictionColor}{1.00} & 0.50 & \colorbox{wrongPredictionColor}{0.53} & 0.50 & \colorbox{wrongPredictionColor}{0.68} & 0.73 & \colorbox{correctPredictionColor}{0.99} & 0.27 & \colorbox{wrongPredictionColor}{0.76} & 1.\\
$\datasetSentimentOOD$ & $\modelPhiThreeFiveInstruct$ & $\estimatorSoftmax$ & \colorbox{correctPredictionColor}{1.00} & 0.50 & \colorbox{wrongPredictionColor}{0.54} & 0.46 & \colorbox{wrongPredictionColor}{0.70} & 0.71 & \colorbox{correctPredictionColor}{0.99} & 0.25 & \colorbox{wrongPredictionColor}{0.78} & 0.96\\
$\datasetSentimentOOD$ & $\modelPhiThreeFiveInstruct$ & $\estimatorTempScaling$ & \colorbox{correctPredictionColor}{1.00} & 0.49 & \colorbox{wrongPredictionColor}{0.58} & 0.30 & \colorbox{wrongPredictionColor}{0.80} & 0.62 & \colorbox{correctPredictionColor}{0.99} & 0.17 & \colorbox{wrongPredictionColor}{0.84} & 0.79\\
$\datasetSentimentOOD$ & $\modelPhiThreeFiveInstruct$ & $\conformalAPS$ & \colorbox{correctPredictionColor}{1.00} & 0.49 & \colorbox{wrongPredictionColor}{0.59} & 0.28 & \colorbox{wrongPredictionColor}{0.81} & 0.60 & \colorbox{correctPredictionColor}{0.99} & 0.17 & \colorbox{wrongPredictionColor}{0.85} & 0.77\\
 
$\datasetSentimentOOD$ & $\modelPhiThreeFiveInstruct$ & $\conformalRAPS$ & \colorbox{correctPredictionColor}{1.00} & 0.49 & \colorbox{wrongPredictionColor}{0.59} & 0.28 & \colorbox{wrongPredictionColor}{0.81} & 0.60 & \colorbox{correctPredictionColor}{0.99} & 0.17 & \colorbox{wrongPredictionColor}{0.85} & 0.77\\
 
$\datasetSentimentOOD$ & $\modelPhiThreeFiveInstructCNNAdaptor$ & $\estimatorNoReject$ & \colorbox{wrongPredictionColor}{0.47} & 0.50 & \colorbox{wrongPredictionColor}{0.70} & 0.50 & \colorbox{wrongPredictionColor}{0.61} & 0.38 & \colorbox{wrongPredictionColor}{0.57} & 0.62 & \colorbox{wrongPredictionColor}{0.59} & 1.\\
$\datasetSentimentOOD$ & $\modelPhiThreeFiveInstructCNNAdaptor$ & $\estimatorSoftmax$ & \colorbox{wrongPredictionColor}{0.57} & 0.03 & \colorbox{correctPredictionColor}{0.96} & 0.07 & \colorbox{wrongPredictionColor}{0.84} & 0.02 & \colorbox{wrongPredictionColor}{0.85} & 0.07 & \colorbox{wrongPredictionColor}{0.85} & 0.09\\
$\datasetSentimentOOD$ & $\modelPhiThreeFiveInstructCNNAdaptor$ & $\estimatorTempScaling$ & \colorbox{wrongPredictionColor}{0.60} & 0.02 & \colorbox{correctPredictionColor}{0.97} & 0.05 & \colorbox{wrongPredictionColor}{0.86} & 0.01 & \colorbox{wrongPredictionColor}{0.87} & 0.06 & \colorbox{wrongPredictionColor}{0.87} & 0.07\\
$\datasetSentimentOOD$ & $\modelPhiThreeFiveInstructCNNAdaptor$ & $\conformalAPS$ & \colorbox{wrongPredictionColor}{0.46} & 0.14 & \colorbox{wrongPredictionColor}{0.83} & 0.18 & \colorbox{wrongPredictionColor}{0.67} & 0.09 & \colorbox{wrongPredictionColor}{0.68} & 0.22 & \colorbox{wrongPredictionColor}{0.68} & 0.32\\
 
$\datasetSentimentOOD$ & $\modelPhiThreeFiveInstructCNNAdaptor$ & $\conformalRAPS$ & \colorbox{wrongPredictionColor}{0.48} & 0.13 & \colorbox{wrongPredictionColor}{0.82} & 0.18 & \colorbox{wrongPredictionColor}{0.66} & 0.10 & \colorbox{wrongPredictionColor}{0.68} & 0.22 & \colorbox{wrongPredictionColor}{0.68} & 0.32\\
 
$\datasetSentimentOOD$ & $\modelPhiThreeFiveInstructSDM$ & $\estimatorNoReject$ & \colorbox{wrongPredictionColor}{0.92} & 0.50 & \colorbox{wrongPredictionColor}{0.84} & 0.50 & \colorbox{wrongPredictionColor}{0.85} & 0.54 & \colorbox{wrongPredictionColor}{0.91} & 0.46 & \colorbox{wrongPredictionColor}{0.88} & 1.\\
$\datasetSentimentOOD$ & $\modelPhiThreeFiveInstructSDM$ & $\estimatorSoftmax$ & \colorbox{correctPredictionColor}{0.96} & 0.42 & \colorbox{wrongPredictionColor}{0.87} & 0.45 & \colorbox{wrongPredictionColor}{0.87} & 0.46 & \colorbox{correctPredictionColor}{0.96} & 0.41 & \colorbox{wrongPredictionColor}{0.91} & 0.87\\
$\datasetSentimentOOD$ & $\modelPhiThreeFiveInstructSDM$ & $\estimatorSoftmaxOverDistanceMagnitude$ & \colorbox{correctPredictionColor}{1.} & <0.01 & \colorbox{correctPredictionColor}{1.} & <0.01 & \colorbox{correctPredictionColor}{1.} & <0.01 & \colorbox{correctPredictionColor}{1.} & <0.01 & \colorbox{correctPredictionColor}{1.} & <0.01\\
$\datasetSentimentOOD$ & $\modelPhiThreeFiveInstructSDM$ & $\sdmAtAlpha$ & \colorbox{correctPredictionColor}{1.} & 0.01 & \colorbox{correctPredictionColor}{0.98} & 0.01 & \colorbox{correctPredictionColor}{0.98} & 0.01 & \colorbox{correctPredictionColor}{1.} & 0.01 & \colorbox{correctPredictionColor}{0.99} & 0.02\\
$\datasetSentimentOOD$ & $\modelPhiThreeFiveInstructSDM$ & $\sdmHR$ & \colorbox{correctPredictionColor}{1.} & <0.01 & \colorbox{correctPredictionColor}{1.} & <0.01 & \colorbox{correctPredictionColor}{1.} & <0.01 & \colorbox{correctPredictionColor}{1.} & <0.01 & \colorbox{correctPredictionColor}{1.} & 0.01\\

  \midrule 
 $\datasetSentimentOOD$ & $\modelMixtral$ & $\estimatorNoReject$ & \colorbox{correctPredictionColor}{1.00} & 0.50 & \colorbox{wrongPredictionColor}{0.35} & 0.50 & \colorbox{wrongPredictionColor}{0.61} & 0.82 & \colorbox{correctPredictionColor}{1.00} & 0.18 & \colorbox{wrongPredictionColor}{0.67} & 1.\\
$\datasetSentimentOOD$ & $\modelMixtral$ & $\estimatorSoftmax$ & \colorbox{correctPredictionColor}{1.00} & 0.50 & \colorbox{wrongPredictionColor}{0.35} & 0.49 & \colorbox{wrongPredictionColor}{0.61} & 0.82 & \colorbox{correctPredictionColor}{1.00} & 0.17 & \colorbox{wrongPredictionColor}{0.68} & 0.99\\
$\datasetSentimentOOD$ & $\modelMixtral$ & $\estimatorTempScaling$ & \colorbox{correctPredictionColor}{1.00} & 0.49 & \colorbox{wrongPredictionColor}{0.37} & 0.41 & \colorbox{wrongPredictionColor}{0.66} & 0.75 & \colorbox{correctPredictionColor}{0.99} & 0.15 & \colorbox{wrongPredictionColor}{0.71} & 0.90\\
$\datasetSentimentOOD$ & $\modelMixtral$ & $\conformalAPS$ & \colorbox{correctPredictionColor}{1.00} & 0.45 & \colorbox{wrongPredictionColor}{0.44} & 0.32 & \colorbox{wrongPredictionColor}{0.71} & 0.63 & \colorbox{correctPredictionColor}{0.99} & 0.14 & \colorbox{wrongPredictionColor}{0.77} & 0.77\\
$\datasetSentimentOOD$ & $\modelMixtral$ & $\conformalRAPS$ & \colorbox{correctPredictionColor}{1.00} & 0.45 & \colorbox{wrongPredictionColor}{0.44} & 0.32 & \colorbox{wrongPredictionColor}{0.72} & 0.63 & \colorbox{correctPredictionColor}{0.99} & 0.14 & \colorbox{wrongPredictionColor}{0.77} & 0.77\\

$\datasetSentimentOOD$ & $\modelMixtralCNNAdaptor$ & $\estimatorNoReject$ & \colorbox{wrongPredictionColor}{0.88} & 0.50 & \colorbox{wrongPredictionColor}{0.51} & 0.50 & \colorbox{wrongPredictionColor}{0.64} & 0.69 & \colorbox{wrongPredictionColor}{0.82} & 0.31 & \colorbox{wrongPredictionColor}{0.70} & 1.\\
$\datasetSentimentOOD$ & $\modelMixtralCNNAdaptor$ & $\estimatorSoftmax$ & \colorbox{correctPredictionColor}{0.98} & 0.02 & \colorbox{wrongPredictionColor}{0.83} & 0.07 & \colorbox{wrongPredictionColor}{0.66} & 0.04 & \colorbox{correctPredictionColor}{0.99} & 0.06 & \colorbox{wrongPredictionColor}{0.87} & 0.10\\
$\datasetSentimentOOD$ & $\modelMixtralCNNAdaptor$ & $\estimatorTempScaling$ & \colorbox{correctPredictionColor}{0.98} & 0.01 & \colorbox{wrongPredictionColor}{0.90} & 0.05 & \colorbox{wrongPredictionColor}{0.67} & 0.02 & \colorbox{correctPredictionColor}{1.00} & 0.05 & \colorbox{wrongPredictionColor}{0.91} & 0.06\\
$\datasetSentimentOOD$ & $\modelMixtralCNNAdaptor$ & $\conformalAPS$ & \colorbox{wrongPredictionColor}{0.94} & 0.14 & \colorbox{wrongPredictionColor}{0.63} & 0.18 & \colorbox{wrongPredictionColor}{0.67} & 0.20 & \colorbox{wrongPredictionColor}{0.93} & 0.12 & \colorbox{wrongPredictionColor}{0.77} & 0.32\\
 
$\datasetSentimentOOD$ & $\modelMixtralCNNAdaptor$ & $\conformalRAPS$ & \colorbox{wrongPredictionColor}{0.94} & 0.14 & \colorbox{wrongPredictionColor}{0.63} & 0.18 & \colorbox{wrongPredictionColor}{0.67} & 0.20 & \colorbox{wrongPredictionColor}{0.93} & 0.12 & \colorbox{wrongPredictionColor}{0.76} & 0.32\\

$\datasetSentimentOOD$ & $\modelMixtralSDM$ & $\estimatorNoReject$ & \colorbox{wrongPredictionColor}{0.71} & 0.50 & \colorbox{wrongPredictionColor}{0.83} & 0.50 & \colorbox{wrongPredictionColor}{0.81} & 0.44 & \colorbox{wrongPredictionColor}{0.74} & 0.56 & \colorbox{wrongPredictionColor}{0.77} & 1.\\
$\datasetSentimentOOD$ & $\modelMixtralSDM$ & $\estimatorSoftmax$ & \colorbox{wrongPredictionColor}{0.74} & 0.43 & \colorbox{wrongPredictionColor}{0.86} & 0.47 & \colorbox{wrongPredictionColor}{0.83} & 0.39 & \colorbox{wrongPredictionColor}{0.78} & 0.52 & \colorbox{wrongPredictionColor}{0.80} & 0.91\\
$\datasetSentimentOOD$ & $\modelMixtralSDM$ & $\estimatorSoftmaxOverDistanceMagnitude$ & \colorbox{correctPredictionColor}{1.} & <0.01 & \colorbox{correctPredictionColor}{0.98} & 0.02 & \colorbox{wrongPredictionColor}{0.78} & <0.01 & \colorbox{correctPredictionColor}{1.} & 0.02 & \colorbox{correctPredictionColor}{0.98} & 0.02\\
$\datasetSentimentOOD$ & $\modelMixtralSDM$ & $\sdmAtAlpha$ & \colorbox{correctPredictionColor}{0.98} & 0.05 & \colorbox{correctPredictionColor}{0.96} & 0.04 & \colorbox{correctPredictionColor}{0.97} & 0.05 & \colorbox{correctPredictionColor}{0.98} & 0.04 & \colorbox{correctPredictionColor}{0.97} & 0.08\\
$\datasetSentimentOOD$ & $\modelMixtralSDM$ & $\sdmHR$ & \colorbox{wrongPredictionColor}{0.9487} & 0.01 & \colorbox{correctPredictionColor}{0.96} & 0.01 & \colorbox{wrongPredictionColor}{0.9487} & 0.01 & \colorbox{correctPredictionColor}{0.96} & 0.01 & \colorbox{correctPredictionColor}{0.95} & 0.02\\

    \bottomrule
  \end{tabular}
  }  
    \caption{Comparison of estimators for the sentiment datasets, with \colorbox{correctPredictionColor}{$\alpha$}$=0.95$. \colorbox{correctPredictionColor}{\allRejected} indicates all predictions were rejected, which is preferred over falling \colorbox{wrongPredictionColor}{under} the expected accuracy. 
  $n=|\text{Admitted}|$, the count of non-rejected documents.} %
  \label{tab:experiments-sentiment} 
\end{table*}

%% file: tables/table-experiments-classification-factcheck.tex
\begin{table*}
  \centering
  \resizebox{1.0\textwidth}{!}{%
  \begin{tabular}{l l  l c c c c c c c c c l }
    \toprule

    & & &  \multicolumn{4}{c}{Class-conditional} & \multicolumn{4}{c}{Prediction-conditional}  & \multicolumn{2}{c}{Marginal} \\
    & & &  \multicolumn{2}{c}{$y=0$} & \multicolumn{2}{c}{$y=1$} & \multicolumn{2}{c}{$\hat{y}=0$} & \multicolumn{2}{c}{$\hat{y}=1$} & \multicolumn{2}{c}{$y\in \{0,1\}$} \\
    \cmidrule(r){4-5} \cmidrule(r){6-7} \cmidrule(r){8-9} \cmidrule(r){10-11} \cmidrule(r){12-13} \\
    Dataset   & Model & Estimator & \textsc{Acc.}& $\frac{n}{|\testSplit|}$ & \textsc{Acc.} & $\frac{n}{|\testSplit|}$ & \textsc{Acc.} & $\frac{n}{|\testSplit|}$ & \textsc{Acc.} & $\frac{n}{|\testSplit|}$ & \textsc{Acc.} & $\frac{n}{|\testSplit|}$\\
  \midrule
 $\datasetFactcheck$ & $\modelPhiThreeFiveInstruct$ & $\estimatorNoReject$ & \colorbox{wrongPredictionColor}{0.94} & 0.51 & \colorbox{wrongPredictionColor}{0.71} & 0.49 & \colorbox{wrongPredictionColor}{0.78} & 0.62 & \colorbox{wrongPredictionColor}{0.92} & 0.38 & \colorbox{wrongPredictionColor}{0.83} & 1.\\
$\datasetFactcheck$ & $\modelPhiThreeFiveInstruct$ & $\estimatorSoftmax$ & \colorbox{wrongPredictionColor}{0.94} & 0.51 & \colorbox{wrongPredictionColor}{0.73} & 0.46 & \colorbox{wrongPredictionColor}{0.79} & 0.60 & \colorbox{wrongPredictionColor}{0.92} & 0.36 & \colorbox{wrongPredictionColor}{0.84} & 0.97\\
$\datasetFactcheck$ & $\modelPhiThreeFiveInstruct$ & $\estimatorTempScaling$ & \colorbox{correctPredictionColor}{0.97} & 0.38 & \colorbox{wrongPredictionColor}{0.79} & 0.37 & \colorbox{wrongPredictionColor}{0.83} & 0.45 & \colorbox{correctPredictionColor}{0.96} & 0.31 & \colorbox{wrongPredictionColor}{0.88} & 0.76\\
$\datasetFactcheck$ & $\modelPhiThreeFiveInstruct$ & $\conformalAPS$ & \colorbox{correctPredictionColor}{0.98} & 0.22 & \colorbox{wrongPredictionColor}{0.82} & 0.27 & \colorbox{wrongPredictionColor}{0.82} & 0.27 & \colorbox{correctPredictionColor}{0.98} & 0.23 & \colorbox{wrongPredictionColor}{0.89} & 0.50\\

$\datasetFactcheck$ & $\modelPhiThreeFiveInstruct$ & $\conformalRAPS$ & \colorbox{correctPredictionColor}{0.98} & 0.20 & \colorbox{wrongPredictionColor}{0.84} & 0.28 & \colorbox{wrongPredictionColor}{0.81} & 0.24 & \colorbox{correctPredictionColor}{0.98} & 0.24 & \colorbox{wrongPredictionColor}{0.90} & 0.47\\
 
$\datasetFactcheck$ & $\modelPhiThreeFiveInstructCNNAdaptor$ & $\estimatorNoReject$ & \colorbox{wrongPredictionColor}{0.33} & 0.51 & \colorbox{wrongPredictionColor}{0.94} & 0.49 & \colorbox{wrongPredictionColor}{0.85} & 0.20 & \colorbox{wrongPredictionColor}{0.57} & 0.80 & \colorbox{wrongPredictionColor}{0.62} & 1.\\
$\datasetFactcheck$ & $\modelPhiThreeFiveInstructCNNAdaptor$ & $\estimatorSoftmax$ & \colorbox{wrongPredictionColor}{0.40} & 0.08 & \colorbox{correctPredictionColor}{0.99} & 0.33 & \colorbox{wrongPredictionColor}{0.89} & 0.04 & \colorbox{wrongPredictionColor}{0.87} & 0.37 & \colorbox{wrongPredictionColor}{0.87} & 0.41\\
$\datasetFactcheck$ & $\modelPhiThreeFiveInstructCNNAdaptor$ & $\estimatorTempScaling$ & \colorbox{wrongPredictionColor}{0.38} & 0.07 & \colorbox{correctPredictionColor}{0.99} & 0.29 & \colorbox{wrongPredictionColor}{0.86} & 0.03 & \colorbox{wrongPredictionColor}{0.88} & 0.33 & \colorbox{wrongPredictionColor}{0.88} & 0.36\\
$\datasetFactcheck$ & $\modelPhiThreeFiveInstructCNNAdaptor$ & $\conformalAPS$ & \colorbox{wrongPredictionColor}{0.26} & 0.14 & \colorbox{correctPredictionColor}{0.99} & 0.38 & \colorbox{wrongPredictionColor}{0.90} & 0.04 & \colorbox{wrongPredictionColor}{0.78} & 0.48 & \colorbox{wrongPredictionColor}{0.79} & 0.52\\
 
$\datasetFactcheck$ & $\modelPhiThreeFiveInstructCNNAdaptor$ & $\conformalRAPS$ & \colorbox{wrongPredictionColor}{0.36} & 0.18 & \colorbox{correctPredictionColor}{0.98} & 0.35 & \colorbox{wrongPredictionColor}{0.89} & 0.07 & \colorbox{wrongPredictionColor}{0.74} & 0.46 & \colorbox{wrongPredictionColor}{0.76} & 0.53\\
 
  $\datasetFactcheck$ & $\modelPhiThreeFiveInstructSDM$ & $\estimatorNoReject$ & \colorbox{wrongPredictionColor}{0.70} & 0.51 & \colorbox{wrongPredictionColor}{0.88} & 0.49 & \colorbox{wrongPredictionColor}{0.86} & 0.42 & \colorbox{wrongPredictionColor}{0.73} & 0.58 & \colorbox{wrongPredictionColor}{0.79} & 1.\\
$\datasetFactcheck$ & $\modelPhiThreeFiveInstructSDM$ & $\estimatorSoftmax$ & \colorbox{wrongPredictionColor}{0.75} & 0.27 & \colorbox{wrongPredictionColor}{0.94} & 0.39 & \colorbox{wrongPredictionColor}{0.89} & 0.22 & \colorbox{wrongPredictionColor}{0.
85} & 0.43 & \colorbox{wrongPredictionColor}{0.86} & 0.65\\
$\datasetFactcheck$ & $\modelPhiThreeFiveInstructSDM$ & $\estimatorSoftmaxOverDistanceMagnitude$ & \colorbox{correctPredictionColor}{\allRejected} & 0. & \colorbox{correctPredictionColor}{1.} & 0.03 & \colorbox{correctPredictionColor}{\allRejected} & 0. & \colorbox{correctPredictionColor}{1.} & 0.03 & \colorbox{correctPredictionColor}{1.} & 0.03\\
$\datasetFactcheck$ & $\modelPhiThreeFiveInstructSDM$ & $\sdmAtAlpha$ & \colorbox{correctPredictionColor}{1.} & 0.01 & \colorbox{correctPredictionColor}{0.97} & 0.14 & \colorbox{wrongPredictionColor}{0.75} & 0.02 & \colorbox{correctPredictionColor}{1.} & 0.14 & \colorbox{correctPredictionColor}{0.97} & 0.16\\
$\datasetFactcheck$ & $\modelPhiThreeFiveInstructSDM$ & $\sdmHR$ & \colorbox{correctPredictionColor}{\allRejected} & 0. & \colorbox{correctPredictionColor}{1.} & 0.12 & \colorbox{correctPredictionColor}{\allRejected} & 0. & \colorbox{correctPredictionColor}{1.} & 0.12 & \colorbox{correctPredictionColor}{1.} & 0.12\\

\midrule 
$\datasetFactcheck$ & $\modelMixtral$ & $\estimatorNoReject$ & \colorbox{correctPredictionColor}{0.98} & 0.51 & \colorbox{wrongPredictionColor}{0.48} & 0.49 & \colorbox{wrongPredictionColor}{0.66} & 0.76 & \colorbox{correctPredictionColor}{0.95} & 0.24 & \colorbox{wrongPredictionColor}{0.73} & 1.\\
$\datasetFactcheck$ & $\modelMixtral$ & $\estimatorSoftmax$ & \colorbox{correctPredictionColor}{0.98} & 0.51 & \colorbox{wrongPredictionColor}{0.48} & 0.49 & \colorbox{wrongPredictionColor}{0.66} & 0.76 & \colorbox{correctPredictionColor}{0.95} & 0.24 & \colorbox{wrongPredictionColor}{0.73} & 1.\\
$\datasetFactcheck$ & $\modelMixtral$ & $\estimatorTempScaling$ & \colorbox{correctPredictionColor}{0.99} & 0.50 & \colorbox{wrongPredictionColor}{0.46} & 0.43 & \colorbox{wrongPredictionColor}{0.68} & 0.73 & \colorbox{correctPredictionColor}{0.98} & 0.20 & \colorbox{wrongPredictionColor}{0.75} & 0.93\\
$\datasetFactcheck$ & $\modelMixtral$ & $\conformalAPS$ & \colorbox{correctPredictionColor}{1.} & 0.18 & \colorbox{wrongPredictionColor}{0.80} & 0.16 & \colorbox{wrongPredictionColor}{0.84} & 0.21 & \colorbox{correctPredictionColor}{1.} & 0.13 & \colorbox{wrongPredictionColor}{0.90} & 0.34\\
 
$\datasetFactcheck$ & $\modelMixtral$ & $\conformalRAPS$ & \colorbox{correctPredictionColor}{1.} & 0.14 & \colorbox{wrongPredictionColor}{0.66} & 0.20 & \colorbox{wrongPredictionColor}{0.67} & 0.21 & \colorbox{correctPredictionColor}{1.} & 0.13 & \colorbox{wrongPredictionColor}{0.80} & 0.35\\
 
$\datasetFactcheck$ & $\modelMixtralCNNAdaptor$ & $\estimatorNoReject$ & \colorbox{wrongPredictionColor}{0.56} & 0.51 & \colorbox{wrongPredictionColor}{0.87} & 0.49 & \colorbox{wrongPredictionColor}{0.82} & 0.36 & \colorbox{wrongPredictionColor}{0.65} & 0.64 & \colorbox{wrongPredictionColor}{0.71} & 1.\\
$\datasetFactcheck$ & $\modelMixtralCNNAdaptor$ & $\estimatorSoftmax$ & \colorbox{wrongPredictionColor}{0.68} & 0.11 & \colorbox{correctPredictionColor}{0.97} & 0.31 & \colorbox{wrongPredictionColor}{0.90} & 0.09 & \colorbox{wrongPredictionColor}{0.89} & 0.34 & \colorbox{wrongPredictionColor}{0.89} & 0.42\\
$\datasetFactcheck$ & $\modelMixtralCNNAdaptor$ & $\estimatorTempScaling$ & \colorbox{wrongPredictionColor}{0.70} & 0.09 & \colorbox{correctPredictionColor}{0.97} & 0.29 & \colorbox{wrongPredictionColor}{0.89} & 0.07 & \colorbox{wrongPredictionColor}{0.91} & 0.31 & \colorbox{wrongPredictionColor}{0.91} & 0.39\\
$\datasetFactcheck$ & $\modelMixtralCNNAdaptor$ & $\conformalAPS$ & \colorbox{wrongPredictionColor}{0.62} & 0.22 & \colorbox{correctPredictionColor}{0.96} & 0.37 & \colorbox{wrongPredictionColor}{0.89} & 0.16 & \colorbox{wrongPredictionColor}{0.80} & 0.44 & \colorbox{wrongPredictionColor}{0.83} & 0.59\\
 
$\datasetFactcheck$ & $\modelMixtralCNNAdaptor$ & $\conformalRAPS$ & \colorbox{wrongPredictionColor}{0.65} & 0.22 & \colorbox{correctPredictionColor}{0.96} & 0.35 & \colorbox{wrongPredictionColor}{0.92} & 0.16 & \colorbox{wrongPredictionColor}{0.81} & 0.41 & \colorbox{wrongPredictionColor}{0.84} & 0.57\\
 
 $\datasetFactcheck$ & $\modelMixtralSDM$ & $\estimatorNoReject$ & \colorbox{wrongPredictionColor}{0.63} & 0.51 & \colorbox{wrongPredictionColor}{0.90} & 0.49 & \colorbox{wrongPredictionColor}{0.87} & 0.38 & \colorbox{wrongPredictionColor}{0.70} & 0.62 & \colorbox{wrongPredictionColor}{0.76} & 1.\\
$\datasetFactcheck$ & $\modelMixtralSDM$ & $\estimatorSoftmax$ & \colorbox{wrongPredictionColor}{0.67} & 0.34 & \colorbox{correctPredictionColor}{0.96} & 0.40 & \colorbox{wrongPredictionColor}{0.93} & 0.24 & \colorbox{wrongPredictionColor}{0.78} & 0.50 & \colorbox{wrongPredictionColor}{0.83} & 0.74\\
$\datasetFactcheck$ & $\modelMixtralSDM$ & $\estimatorSoftmaxOverDistanceMagnitude$ & \colorbox{correctPredictionColor}{\allRejected} & 0. & \colorbox{wrongPredictionColor}{0.80} & 0.04 & \colorbox{wrongPredictionColor}{0.} & 0.01 & \colorbox{correctPredictionColor}{1.} & 0.03 & \colorbox{wrongPredictionColor}{0.80} & 0.04\\
$\datasetFactcheck$ & $\modelMixtralSDM$ & $\sdmAtAlpha$ & \colorbox{wrongPredictionColor}{0.88} & 0.10 & \colorbox{correctPredictionColor}{0.95} & 0.18 & \colorbox{wrongPredictionColor}{0.91} & 0.09 & \colorbox{wrongPredictionColor}{0.93} & 0.18 & \colorbox{wrongPredictionColor}{0.93} & 0.27\\
$\datasetFactcheck$ & $\modelMixtralSDM$ & $\sdmHR$ & \colorbox{correctPredictionColor}{\allRejected} & 0. & \colorbox{correctPredictionColor}{\allRejected} & 0. & \colorbox{correctPredictionColor}{\allRejected} & 0. & \colorbox{correctPredictionColor}{\allRejected} & 0. & \colorbox{correctPredictionColor}{\allRejected} & 0.\\

$\datasetFactcheck$ & $\modelMixtralSDM$ & $\sdmAtAlpha, \alpha=0.94$ & \colorbox{wrongPredictionColor}{0.85} & 0.11 & \colorbox{correctPredictionColor}{0.95} & 0.18 & \colorbox{wrongPredictionColor}{0.92} & 0.10 & \colorbox{wrongPredictionColor}{0.91} & 0.19 & \colorbox{wrongPredictionColor}{0.91} & 0.29\\
$\datasetFactcheck$ & $\modelMixtralSDM$ & $\sdmHR, \alpha=0.94$ & \colorbox{correctPredictionColor}{1.} & 0.03 & \colorbox{correctPredictionColor}{0.95} & 0.16 & \colorbox{wrongPredictionColor}{0.80} & 0.04 & \colorbox{correctPredictionColor}{1.} & 0.15 & \colorbox{correctPredictionColor}{0.96} & 0.19\\

    \bottomrule
  \end{tabular}
  }  
    \caption{Comparison of estimators for the factcheck datasets. Unless specified otherwise, \colorbox{correctPredictionColor}{$\alpha$}$=0.95$. \colorbox{correctPredictionColor}{\allRejected} indicates all predictions were rejected, which is preferred over falling \colorbox{wrongPredictionColor}{under} the expected accuracy. 
  $n=|\text{Admitted}|$, the count of non-rejected documents.} %
  \label{tab:experiments-factcheck} 
\end{table*}

%% file: tables/table-experiments-classification-sentiment-shuffled.tex
\begin{table*}

  \centering
  \resizebox{1.0\textwidth}{!}{%
\begin{tabular}{l l  l c c c c c c c c c l }
    \toprule

    & & &  \multicolumn{4}{c}{Class-conditional} & \multicolumn{4}{c}{Prediction-conditional}  & \multicolumn{2}{c}{Marginal} \\
    & & &  \multicolumn{2}{c}{$y=0$} & \multicolumn{2}{c}{$y=1$} & \multicolumn{2}{c}{$\hat{y}=0$} & \multicolumn{2}{c}{$\hat{y}=1$} & \multicolumn{2}{c}{$y\in \{0,1\}$} \\
    \cmidrule(r){4-5} \cmidrule(r){6-7} \cmidrule(r){8-9} \cmidrule(r){10-11} \cmidrule(r){12-13} \\
    Dataset   & Model & Estimator & \textsc{Acc.}& $\frac{n}{|\testSplit|}$ & \textsc{Acc.} & $\frac{n}{|\testSplit|}$ & \textsc{Acc.} & $\frac{n}{|\testSplit|}$ & \textsc{Acc.} & $\frac{n}{|\testSplit|}$ & \textsc{Acc.} & $\frac{n}{|\testSplit|}$\\
  \midrule 
$\datasetSentimentShuffled$ & $\modelPhiThreeFiveInstruct$ & $\estimatorNoReject$ & \colorbox{correctPredictionColor}{1.00} & 0.50 & \colorbox{wrongPredictionColor}{0.18} & 0.50 & \colorbox{wrongPredictionColor}{0.55} & 0.91 & \colorbox{correctPredictionColor}{0.98} & 0.09 & \colorbox{wrongPredictionColor}{0.59} & 1.\\
$\datasetSentimentShuffled$ & $\modelPhiThreeFiveInstruct$ & $\estimatorSoftmax$ & \colorbox{correctPredictionColor}{1.00} & 0.50 & \colorbox{wrongPredictionColor}{0.16} & 0.45 & \colorbox{wrongPredictionColor}{0.57} & 0.88 & \colorbox{correctPredictionColor}{0.99} & 0.07 & \colorbox{wrongPredictionColor}{0.60} & 0.95\\
$\datasetSentimentShuffled$ & $\modelPhiThreeFiveInstruct$ & $\estimatorTempScaling$ & \colorbox{correctPredictionColor}{1.} & 0.44 & \colorbox{wrongPredictionColor}{0.12} & 0.18 & \colorbox{wrongPredictionColor}{0.74} & 0.60 & \colorbox{correctPredictionColor}{1.} & 0.02 & \colorbox{wrongPredictionColor}{0.75} & 0.62\\
$\datasetSentimentShuffled$ & $\modelPhiThreeFiveInstruct$ & $\conformalAPS$ & \colorbox{correctPredictionColor}{1.00} & 0.42 & \colorbox{wrongPredictionColor}{0.14} & 0.15 & \colorbox{wrongPredictionColor}{0.76} & 0.55 & \colorbox{correctPredictionColor}{0.97} & 0.02 & \colorbox{wrongPredictionColor}{0.77} & 0.57\\
\bottomrule
  \end{tabular}
  }  
    \caption{Comparison of estimators for the \underline{shuffled} sentiment datasets, with \colorbox{correctPredictionColor}{$\alpha$}$=0.95$. \colorbox{correctPredictionColor}{\allRejected} indicates all predictions were rejected, which is preferred over falling \colorbox{wrongPredictionColor}{under} the expected accuracy. 
  $n=|\text{Admitted}|$, the count of non-rejected documents.} %
  \label{tab:experiments-sentiment-shuffled} 
\end{table*}

%% file: tables/table-experiments-classification-factcheck-shuffled.tex
\begin{table*}
  \centering
  \resizebox{1.0\textwidth}{!}{%
  \begin{tabular}{l l  l c c c c c c c c c l }
    \toprule

    & & &  \multicolumn{4}{c}{Class-conditional} & \multicolumn{4}{c}{Prediction-conditional}  & \multicolumn{2}{c}{Marginal} \\
    & & &  \multicolumn{2}{c}{$y=0$} & \multicolumn{2}{c}{$y=1$} & \multicolumn{2}{c}{$\hat{y}=0$} & \multicolumn{2}{c}{$\hat{y}=1$} & \multicolumn{2}{c}{$y\in \{0,1\}$} \\
    \cmidrule(r){4-5} \cmidrule(r){6-7} \cmidrule(r){8-9} \cmidrule(r){10-11} \cmidrule(r){12-13} \\
    Dataset   & Model & Estimator & \textsc{Acc.}& $\frac{n}{|\testSplit|}$ & \textsc{Acc.} & $\frac{n}{|\testSplit|}$ & \textsc{Acc.} & $\frac{n}{|\testSplit|}$ & \textsc{Acc.} & $\frac{n}{|\testSplit|}$ & \textsc{Acc.} & $\frac{n}{|\testSplit|}$\\
  \midrule

$\datasetFactcheckShuffled$ & $\modelPhiThreeFiveInstruct$ & $\estimatorNoReject$ & \colorbox{wrongPredictionColor}{0.91} & 1. & - & 0. & \colorbox{correctPredictionColor}{1.} & 0.91 & \colorbox{wrongPredictionColor}{0.} & 0.09 & \colorbox{wrongPredictionColor}{0.91} & 1.\\
$\datasetFactcheckShuffled$ & $\modelPhiThreeFiveInstruct$ & $\estimatorSoftmax$ & \colorbox{wrongPredictionColor}{0.92} & 0.99 & - & 0. & \colorbox{correctPredictionColor}{1.} & 0.91 & \colorbox{wrongPredictionColor}{0.} & 0.08 & \colorbox{wrongPredictionColor}{0.92} & 0.99\\
$\datasetFactcheckShuffled$ & $\modelPhiThreeFiveInstruct$ & $\estimatorTempScaling$ & \colorbox{wrongPredictionColor}{0.93} & 0.87 & - & 0. & \colorbox{correctPredictionColor}{1.} & 0.81 & \colorbox{wrongPredictionColor}{0.} & 0.06 & \colorbox{wrongPredictionColor}{0.93} & 0.87\\
$\datasetFactcheckShuffled$ & $\modelPhiThreeFiveInstruct$ & $\conformalAPS$ & \colorbox{wrongPredictionColor}{0.93} & 0.45 & - & 0. & \colorbox{correctPredictionColor}{1.} & 0.42 & \colorbox{wrongPredictionColor}{0.} & 0.03 & \colorbox{wrongPredictionColor}{0.93} & 0.45\\
 
$\datasetFactcheckShuffled$ & $\modelPhiThreeFiveInstruct$ & $\conformalRAPS$ & \colorbox{correctPredictionColor}{0.95} & 0.52 & - & 0. & \colorbox{correctPredictionColor}{1.} & 0.50 & \colorbox{wrongPredictionColor}{0.} & 0.02 & \colorbox{correctPredictionColor}{0.95} & 0.52\\
 
$\datasetFactcheckShuffled$ & $\modelPhiThreeFiveInstructCNNAdaptor$ & $\estimatorNoReject$ & \colorbox{wrongPredictionColor}{0.34} & 1. & - & 0. & \colorbox{correctPredictionColor}{1.} & 0.34 & \colorbox{wrongPredictionColor}{0.} & 0.66 & \colorbox{wrongPredictionColor}{0.34} & 1.\\
$\datasetFactcheckShuffled$ & $\modelPhiThreeFiveInstructCNNAdaptor$ & $\estimatorSoftmax$ & \colorbox{wrongPredictionColor}{0.20} & 0.24 & - & 0. & \colorbox{correctPredictionColor}{1.} & 0.05 & \colorbox{wrongPredictionColor}{0.} & 0.19 & \colorbox{wrongPredictionColor}{0.20} & 0.24\\
$\datasetFactcheckShuffled$ & $\modelPhiThreeFiveInstructCNNAdaptor$ & $\estimatorTempScaling$ & \colorbox{wrongPredictionColor}{0.13} & 0.19 &- & 0. & \colorbox{correctPredictionColor}{1.} & 0.02 & \colorbox{wrongPredictionColor}{0.} & 0.17 & \colorbox{wrongPredictionColor}{0.13} & 0.19\\
$\datasetFactcheckShuffled$ & $\modelPhiThreeFiveInstructCNNAdaptor$ & $\conformalAPS$ & \colorbox{wrongPredictionColor}{0.24} & 0.38 & - & 0. & \colorbox{correctPredictionColor}{1.} & 0.09 & \colorbox{wrongPredictionColor}{0.} & 0.29 & \colorbox{wrongPredictionColor}{0.24} & 0.38\\
 
$\datasetFactcheckShuffled$ & $\modelPhiThreeFiveInstructCNNAdaptor$ & $\conformalRAPS$ & \colorbox{wrongPredictionColor}{0.27} & 0.39 & - & 0. & \colorbox{correctPredictionColor}{1.} & 0.11 & \colorbox{wrongPredictionColor}{0.} & 0.29 & \colorbox{wrongPredictionColor}{0.27} & 0.39\\
 
$\datasetFactcheckShuffled$ & $\modelPhiThreeFiveInstructSDM$ & $\estimatorNoReject$ & \colorbox{wrongPredictionColor}{0.66} & 1. & -& 0. & \colorbox{correctPredictionColor}{1.} & 0.66 & \colorbox{wrongPredictionColor}{0.} & 0.34 & \colorbox{wrongPredictionColor}{0.66} & 1.\\
$\datasetFactcheckShuffled$ & $\modelPhiThreeFiveInstructSDM$ & $\estimatorSoftmax$ & \colorbox{wrongPredictionColor}{0.69} & 0.64 & - & 0. & \colorbox{correctPredictionColor}{1.} & 0.44 & \colorbox{wrongPredictionColor}{0.} & 0.20 & \colorbox{wrongPredictionColor}{0.69} & 0.64\\
$\datasetFactcheckShuffled$ & $\modelPhiThreeFiveInstructSDM$ & $\estimatorSoftmaxOverDistanceMagnitude$ & \colorbox{correctPredictionColor}{\allRejected} & 0. & - & 0. & \colorbox{correctPredictionColor}{\allRejected} & 0. & \colorbox{correctPredictionColor}{\allRejected} & 0. & \colorbox{correctPredictionColor}{\allRejected} & 0.\\
$\datasetFactcheckShuffled$ & $\modelPhiThreeFiveInstructSDM$ & $\sdmAtAlpha$ & \colorbox{correctPredictionColor}{\allRejected} & 0. & - & 0. & \colorbox{correctPredictionColor}{\allRejected} & 0. & \colorbox{correctPredictionColor}{\allRejected} & 0. & \colorbox{correctPredictionColor}{\allRejected} & 0.\\
$\datasetFactcheckShuffled$ & $\modelPhiThreeFiveInstructSDM$ & $\sdmHR$ & \colorbox{correctPredictionColor}{\allRejected} & 0. & - & 0. & \colorbox{correctPredictionColor}{\allRejected} & 0. & \colorbox{correctPredictionColor}{\allRejected} & 0. & \colorbox{correctPredictionColor}{\allRejected} & 0.\\
  \midrule 
$\datasetFactcheckShuffled$ & $\modelMixtral$ & $\estimatorNoReject$ & \colorbox{correctPredictionColor}{0.98} & 1. & - & 0. & \colorbox{correctPredictionColor}{1.} & 0.98 & \colorbox{wrongPredictionColor}{0.} & 0.02 & \colorbox{correctPredictionColor}{0.98} & 1.\\
$\datasetFactcheckShuffled$ & $\modelMixtral$ & $\estimatorSoftmax$ & \colorbox{correctPredictionColor}{0.98} & 1. & - & 0. & \colorbox{correctPredictionColor}{1.} & 0.98 & \colorbox{wrongPredictionColor}{0.} & 0.02 & \colorbox{correctPredictionColor}{0.98} & 1.\\
$\datasetFactcheckShuffled$ & $\modelMixtral$ & $\estimatorTempScaling$ & \colorbox{correctPredictionColor}{0.98} & 0.98 & - & 0. & \colorbox{correctPredictionColor}{1.} & 0.96 & \colorbox{wrongPredictionColor}{0.} & 0.02 & \colorbox{correctPredictionColor}{0.98} & 0.98\\
$\datasetFactcheckShuffled$ & $\modelMixtral$ & $\conformalAPS$ & \colorbox{correctPredictionColor}{0.98} & 0.18 & - & 0. & \colorbox{correctPredictionColor}{1.} & 0.18 & \colorbox{wrongPredictionColor}{0.} & <0.01 & \colorbox{correctPredictionColor}{0.98} & 0.18\\
 
$\datasetFactcheckShuffled$ & $\modelMixtral$ & $\conformalRAPS$ & \colorbox{correctPredictionColor}{0.98} & 0.23 & - & 0. & \colorbox{correctPredictionColor}{1.} & 0.23 & \colorbox{wrongPredictionColor}{0.} & <0.01 & \colorbox{correctPredictionColor}{0.98} & 0.23\\
 
$\datasetFactcheckShuffled$ & $\modelMixtralCNNAdaptor$ & $\estimatorNoReject$ & \colorbox{wrongPredictionColor}{0.79} & 1. & -& 0. & \colorbox{correctPredictionColor}{1.} & 0.79 & \colorbox{wrongPredictionColor}{0.} & 0.21 & \colorbox{wrongPredictionColor}{0.79} & 1.\\
$\datasetFactcheckShuffled$ & $\modelMixtralCNNAdaptor$ & $\estimatorSoftmax$ & \colorbox{wrongPredictionColor}{0.69} & 0.13 & - & 0. & \colorbox{correctPredictionColor}{1.} & 0.09 & \colorbox{wrongPredictionColor}{0.} & 0.04 & \colorbox{wrongPredictionColor}{0.69} & 0.13\\
$\datasetFactcheckShuffled$ & $\modelMixtralCNNAdaptor$ & $\estimatorTempScaling$ & \colorbox{wrongPredictionColor}{0.55} & 0.09 & - & 0. & \colorbox{correctPredictionColor}{1.} & 0.05 & \colorbox{wrongPredictionColor}{0.} & 0.04 & \colorbox{wrongPredictionColor}{0.55} & 0.09\\
$\datasetFactcheckShuffled$ & $\modelMixtralCNNAdaptor$ & $\conformalAPS$ & \colorbox{wrongPredictionColor}{0.77} & 0.40 & - & 0. & \colorbox{correctPredictionColor}{1.} & 0.31 & \colorbox{wrongPredictionColor}{0.} & 0.09 & \colorbox{wrongPredictionColor}{0.77} & 0.40\\
 
$\datasetFactcheckShuffled$ & $\modelMixtralCNNAdaptor$ & $\conformalRAPS$ & \colorbox{wrongPredictionColor}{0.79} & 0.39 &- & 0. & \colorbox{correctPredictionColor}{1.} & 0.31 & \colorbox{wrongPredictionColor}{0.} & 0.08 & \colorbox{wrongPredictionColor}{0.79} & 0.39\\
 
 $\datasetFactcheckShuffled$ & $\modelMixtralSDM$ & $\estimatorNoReject$ & \colorbox{wrongPredictionColor}{0.76} & 1. & - & 0. & \colorbox{correctPredictionColor}{1.} & 0.76 & \colorbox{wrongPredictionColor}{0.} & 0.24 & \colorbox{wrongPredictionColor}{0.76} & 1.\\
$\datasetFactcheckShuffled$ & $\modelMixtralSDM$ & $\estimatorSoftmax$ & \colorbox{wrongPredictionColor}{0.79} & 0.65 & - & 0. & \colorbox{correctPredictionColor}{1.} & 0.51 & \colorbox{wrongPredictionColor}{0.} & 0.13 & \colorbox{wrongPredictionColor}{0.79} & 0.65\\
$\datasetFactcheckShuffled$ & $\modelMixtralSDM$ & $\estimatorSoftmaxOverDistanceMagnitude$ & \colorbox{correctPredictionColor}{\allRejected} & 0. & - & 0. & \colorbox{correctPredictionColor}{\allRejected} & 0. & \colorbox{correctPredictionColor}{\allRejected} & 0. & \colorbox{correctPredictionColor}{\allRejected} & 0.\\
$\datasetFactcheckShuffled$ & $\modelMixtralSDM$ & $\sdmAtAlpha$ & \colorbox{correctPredictionColor}{1.} & 0.01 & -& 0. & \colorbox{correctPredictionColor}{1.} & 0.01 & \colorbox{correctPredictionColor}{\allRejected} & 0. & \colorbox{correctPredictionColor}{1.} & 0.01\\
$\datasetFactcheckShuffled$ & $\modelMixtralSDM$ & $\sdmHR$ & \colorbox{correctPredictionColor}{\allRejected} & 0. & - & 0. & \colorbox{correctPredictionColor}{\allRejected} & 0. & \colorbox{correctPredictionColor}{\allRejected} & 0. & \colorbox{correctPredictionColor}{\allRejected} & 0.\\

$\datasetFactcheckShuffled$ & $\modelMixtralSDM$ & $\sdmAtAlpha, \alpha=0.94$ & \colorbox{correctPredictionColor}{1.} & 0.01 & - & 0. & \colorbox{correctPredictionColor}{1.} & 0.01 & \colorbox{correctPredictionColor}{\allRejected} & 0. & \colorbox{correctPredictionColor}{1.} & 0.01\\
$\datasetFactcheckShuffled$ & $\modelMixtralSDM$ & $\sdmHR, \alpha=0.94$ & \colorbox{correctPredictionColor}{1.} & 0.01 & - & 0. & \colorbox{correctPredictionColor}{1.} & 0.01 & \colorbox{correctPredictionColor}{\allRejected} & 0. & \colorbox{correctPredictionColor}{1.} & 0.01\\

    \bottomrule
  \end{tabular}
  }  
    \caption{Comparison of estimators for the \underline{shuffled} factcheck datasets. Unless specified otherwise, \colorbox{correctPredictionColor}{$\alpha$}$=0.95$. \colorbox{correctPredictionColor}{\allRejected} indicates all predictions were rejected, which is preferred over falling \colorbox{wrongPredictionColor}{under} the expected accuracy. 
  $n=|\text{Admitted}|$, the count of non-rejected documents.} %
  \label{tab:experiments-factcheck-shuffled} 
\end{table*}

%% file: tables/additional/table-experiments-classification-sentiment-additional.tex
\begin{table*}
  \centering
  \resizebox{0.8\textwidth}{!}{%
\begin{tabular}{l l  l c c c c c c c c c l }
    \toprule

    & & &  \multicolumn{4}{c}{Class-conditional} & \multicolumn{4}{c}{Prediction-conditional}  & \multicolumn{2}{c}{Marginal} \\
    & & &  \multicolumn{2}{c}{$y=0$} & \multicolumn{2}{c}{$y=1$} & \multicolumn{2}{c}{$\hat{y}=0$} & \multicolumn{2}{c}{$\hat{y}=1$} & \multicolumn{2}{c}{$y\in \{0,1\}$} \\
    \cmidrule(r){4-5} \cmidrule(r){6-7} \cmidrule(r){8-9} \cmidrule(r){10-11} \cmidrule(r){12-13} \\
    Dataset   & Model & Estimator & \textsc{Acc.}& $\frac{n}{|\testSplit|}$ & \textsc{Acc.} & $\frac{n}{|\testSplit|}$ & \textsc{Acc.} & $\frac{n}{|\testSplit|}$ & \textsc{Acc.} & $\frac{n}{|\testSplit|}$ & \textsc{Acc.} & $\frac{n}{|\testSplit|}$\\
  \midrule 
$\datasetSentiment$ $\calibrationSplit$ & $\modelPhiThreeFiveInstructDiscVBLLMLP$ & $\estimatorNoReject$ & \colorbox{correctPredictionColor}{0.97} & 0.51 & \colorbox{correctPredictionColor}{0.96} & 0.49 & \colorbox{correctPredictionColor}{0.96} & 0.51 & \colorbox{correctPredictionColor}{0.96} & 0.49 & \colorbox{correctPredictionColor}{0.96} & 1.\\
$\datasetSentiment$ & $\modelPhiThreeFiveInstructDiscVBLLMLP$ & $\estimatorNoReject$ & \colorbox{correctPredictionColor}{0.97} & 0.50 & \colorbox{correctPredictionColor}{0.95} & 0.50 & \colorbox{correctPredictionColor}{0.95} & 0.51 & \colorbox{correctPredictionColor}{0.97} & 0.49 & \colorbox{correctPredictionColor}{0.96} & 1.\\
$\datasetSentimentOOD$ & $\modelPhiThreeFiveInstructDiscVBLLMLP$ & $\estimatorNoReject$ & \colorbox{wrongPredictionColor}{0.61} & 0.50 & \colorbox{wrongPredictionColor}{0.67} & 0.50 & \colorbox{wrongPredictionColor}{0.65} & 0.47 & \colorbox{wrongPredictionColor}{0.63} & 0.53 & \colorbox{wrongPredictionColor}{0.64} & 1.\\
\bottomrule

  \end{tabular}
  }  
    \caption{Comparison of Bayesian last-layer estimators \citep{HarrisonEtAl-2024-VBLL} for the sentiment datasets, including the shuffled challenge sets, with \colorbox{correctPredictionColor}{$\alpha$}$=0.95$. \colorbox{correctPredictionColor}{\allRejected} indicates all predictions were rejected, which is preferred over falling \colorbox{wrongPredictionColor}{under} the expected accuracy. 
  $n=|\text{Admitted}|$, the count of non-rejected documents.} %
  \label{tab:experiments-sentiment-additional} 
\end{table*}

%% file: tables/additional/table-experiments-classification-factcheck-additional.tex
\begin{table*}
  \centering
  \resizebox{0.8\textwidth}{!}{%
  \begin{tabular}{l l  l c c c c c c c c c l }
    \toprule

    & & &  \multicolumn{4}{c}{Class-conditional} & \multicolumn{4}{c}{Prediction-conditional}  & \multicolumn{2}{c}{Marginal} \\
    & & &  \multicolumn{2}{c}{$y=0$} & \multicolumn{2}{c}{$y=1$} & \multicolumn{2}{c}{$\hat{y}=0$} & \multicolumn{2}{c}{$\hat{y}=1$} & \multicolumn{2}{c}{$y\in \{0,1\}$} \\
    \cmidrule(r){4-5} \cmidrule(r){6-7} \cmidrule(r){8-9} \cmidrule(r){10-11} \cmidrule(r){12-13} \\
    Dataset   & Model & Estimator & \textsc{Acc.}& $\frac{n}{|\testSplit|}$ & \textsc{Acc.} & $\frac{n}{|\testSplit|}$ & \textsc{Acc.} & $\frac{n}{|\testSplit|}$ & \textsc{Acc.} & $\frac{n}{|\testSplit|}$ & \textsc{Acc.} & $\frac{n}{|\testSplit|}$\\
  \midrule
  
  $\datasetFactcheck$ $\calibrationSplit$ & $\modelPhiThreeFiveInstructDiscVBLLMLP$ & $\estimatorNoReject$ & \colorbox{wrongPredictionColor}{0.92} & 0.50 & \colorbox{wrongPredictionColor}{0.91} & 0.50 & \colorbox{wrongPredictionColor}{0.91} & 0.50 & \colorbox{wrongPredictionColor}{0.92} & 0.50 & \colorbox{wrongPredictionColor}{0.91} & 1.\\
    $\datasetFactcheck$ & $\modelPhiThreeFiveInstructDiscVBLLMLP$ & $\estimatorNoReject$ & \colorbox{wrongPredictionColor}{0.38} & 0.51 & \colorbox{wrongPredictionColor}{0.92} & 0.49 & \colorbox{wrongPredictionColor}{0.84} & 0.23 & \colorbox{wrongPredictionColor}{0.59} & 0.77 & \colorbox{wrongPredictionColor}{0.64} & 1.\\
  $\datasetFactcheckShuffled$ & $\modelPhiThreeFiveInstructDiscVBLLMLP$ & $\estimatorNoReject$ & \colorbox{wrongPredictionColor}{0.33} & 1. & \colorbox{correctPredictionColor}{\allRejected} & 0. & \colorbox{correctPredictionColor}{1.} & 0.33 & \colorbox{wrongPredictionColor}{0.} & 0.67 & \colorbox{wrongPredictionColor}{0.33} & 1.\\
  
$\datasetFactcheck$ $\calibrationSplit$ & $\modelPhiThreeFiveInstructDiscVBLLMLP$ & $\estimatorVBLL$ & \colorbox{correctPredictionColor}{0.98} & 0.32 & \colorbox{correctPredictionColor}{0.99} & 0.28 & \colorbox{correctPredictionColor}{0.99} & 0.32 & \colorbox{correctPredictionColor}{0.98} & 0.29 & \colorbox{correctPredictionColor}{0.98} & 0.61\\
$\datasetFactcheck$ & $\modelPhiThreeFiveInstructDiscVBLLMLP$ & $\estimatorVBLL$ & \colorbox{wrongPredictionColor}{0.35} & 0.07 & \colorbox{correctPredictionColor}{1.} & 0.31 & \colorbox{correctPredictionColor}{1.} & 0.02 & \colorbox{wrongPredictionColor}{0.88} & 0.36 & \colorbox{wrongPredictionColor}{0.88} & 0.38\\
$\datasetFactcheckShuffled$ & $\modelPhiThreeFiveInstructDiscVBLLMLP$ & $\estimatorVBLL$ & \colorbox{wrongPredictionColor}{0.15} & 0.21 & \colorbox{correctPredictionColor}{\allRejected} & 0. & \colorbox{correctPredictionColor}{1.} & 0.03 & \colorbox{wrongPredictionColor}{0.} & 0.18 & \colorbox{wrongPredictionColor}{0.15} & 0.21\\
  
  \midrule
  $\datasetFactcheck$ $\calibrationSplit$ & $\modelPhiThreeFiveInstructDiscVBLLMLPrFifty$ & $\estimatorNoReject$ & \colorbox{wrongPredictionColor}{0.91} & 0.50 & \colorbox{wrongPredictionColor}{0.92} & 0.50 & \colorbox{wrongPredictionColor}{0.92} & 0.49 & \colorbox{wrongPredictionColor}{0.91} & 0.51 & \colorbox{wrongPredictionColor}{0.91} & 1.\\
    $\datasetFactcheck$ & $\modelPhiThreeFiveInstructDiscVBLLMLPrFifty$ & $\estimatorNoReject$ & \colorbox{wrongPredictionColor}{0.45} & 0.51 & \colorbox{wrongPredictionColor}{0.94} & 0.49 & \colorbox{wrongPredictionColor}{0.89} & 0.26 & \colorbox{wrongPredictionColor}{0.62} & 0.74 & \colorbox{wrongPredictionColor}{0.69} & 1.\\
  $\datasetFactcheckShuffled$ & $\modelPhiThreeFiveInstructDiscVBLLMLPrFifty$ & $\estimatorNoReject$ & \colorbox{wrongPredictionColor}{0.43} & 1. & \colorbox{correctPredictionColor}{\allRejected} & 0. & \colorbox{correctPredictionColor}{1.} & 0.43 & \colorbox{wrongPredictionColor}{0.} & 0.57 & \colorbox{wrongPredictionColor}{0.43} & 1.\\
  
$\datasetFactcheck$ $\calibrationSplit$ & $\modelPhiThreeFiveInstructDiscVBLLMLPrFifty$ & $\estimatorVBLL$ & \colorbox{correctPredictionColor}{0.98} & 0.34 & \colorbox{correctPredictionColor}{0.98} & 0.33 & \colorbox{correctPredictionColor}{0.98} & 0.34 & \colorbox{correctPredictionColor}{0.98} & 0.33 & \colorbox{correctPredictionColor}{0.98} & 0.67\\
$\datasetFactcheck$ & $\modelPhiThreeFiveInstructDiscVBLLMLPrFifty$ & $\estimatorVBLL$ & \colorbox{wrongPredictionColor}{0.35} & 0.11 & \colorbox{correctPredictionColor}{0.99} & 0.32 & \colorbox{wrongPredictionColor}{0.90} & 0.04 & \colorbox{wrongPredictionColor}{0.82} & 0.39 & \colorbox{wrongPredictionColor}{0.83} & 0.43\\
$\datasetFactcheckShuffled$ & $\modelPhiThreeFiveInstructDiscVBLLMLPrFifty$ & $\estimatorVBLL$ & \colorbox{wrongPredictionColor}{0.33} & 0.33 & \colorbox{correctPredictionColor}{\allRejected} & 0. & \colorbox{correctPredictionColor}{1.} & 0.11 & \colorbox{wrongPredictionColor}{0.} & 0.22 & \colorbox{wrongPredictionColor}{0.33} & 0.33\\

\midrule

$\datasetFactcheck$ $\calibrationSplit$ & $\modelPhiThreeFiveInstructGenVBLLMLP$ & $\estimatorNoReject$ & \colorbox{wrongPredictionColor}{0.90} & 0.50 & \colorbox{wrongPredictionColor}{0.93} & 0.50 & \colorbox{wrongPredictionColor}{0.93} & 0.49 & \colorbox{wrongPredictionColor}{0.91} & 0.51 & \colorbox{wrongPredictionColor}{0.92} & 1.\\
$\datasetFactcheck$ & $\modelPhiThreeFiveInstructGenVBLLMLP$ & $\estimatorNoReject$ & \colorbox{wrongPredictionColor}{0.41} & 0.51 & \colorbox{wrongPredictionColor}{0.93} & 0.49 & \colorbox{wrongPredictionColor}{0.87} & 0.24 & \colorbox{wrongPredictionColor}{0.60} & 0.76 & \colorbox{wrongPredictionColor}{0.67} & 1.\\
$\datasetFactcheckShuffled$ & $\modelPhiThreeFiveInstructGenVBLLMLP$ & $\estimatorNoReject$ & \colorbox{wrongPredictionColor}{0.40} & 1. & \colorbox{correctPredictionColor}{\allRejected} & 0. & \colorbox{correctPredictionColor}{1.} & 0.40 & \colorbox{wrongPredictionColor}{0.} & 0.60 & \colorbox{wrongPredictionColor}{0.40} & 1.\\

$\datasetFactcheck$ $\calibrationSplit$ & $\modelPhiThreeFiveInstructGenVBLLMLP$ & $\estimatorVBLL$ & \colorbox{correctPredictionColor}{0.98} & 0.33 & \colorbox{correctPredictionColor}{0.99} & 0.32 & \colorbox{correctPredictionColor}{0.99} & 0.33 & \colorbox{correctPredictionColor}{0.98} & 0.33 & \colorbox{correctPredictionColor}{0.98} & 0.65\\
$\datasetFactcheck$ & $\modelPhiThreeFiveInstructGenVBLLMLP$ & $\estimatorVBLL$ & \colorbox{wrongPredictionColor}{0.31} & 0.07 & \colorbox{correctPredictionColor}{0.99} & 0.30 & \colorbox{wrongPredictionColor}{0.83} & 0.02 & \colorbox{wrongPredictionColor}{0.87} & 0.34 & \colorbox{wrongPredictionColor}{0.87} & 0.37\\
$\datasetFactcheckShuffled$ & $\modelPhiThreeFiveInstructGenVBLLMLP$ & $\estimatorVBLL$ & \colorbox{wrongPredictionColor}{0.20} & 0.22 & \colorbox{correctPredictionColor}{\allRejected} & 0. & \colorbox{correctPredictionColor}{1.} & 0.04 & \colorbox{wrongPredictionColor}{0.} & 0.18 & \colorbox{wrongPredictionColor}{0.20} & 0.22\\

\midrule
$\datasetFactcheck$ $\calibrationSplit$ & $\modelPhiThreeFiveInstructGenVBLLMLPrFifty$ & $\estimatorNoReject$ & \colorbox{wrongPredictionColor}{0.91} & 0.50 & \colorbox{wrongPredictionColor}{0.93} & 0.50 & \colorbox{wrongPredictionColor}{0.93} & 0.49 & \colorbox{wrongPredictionColor}{0.91} & 0.51 & \colorbox{wrongPredictionColor}{0.92} & 1.\\
$\datasetFactcheck$ & $\modelPhiThreeFiveInstructGenVBLLMLPrFifty$ & $\estimatorNoReject$ & \colorbox{wrongPredictionColor}{0.41} & 0.51 & \colorbox{wrongPredictionColor}{0.93} & 0.49 & \colorbox{wrongPredictionColor}{0.87} & 0.24 & \colorbox{wrongPredictionColor}{0.60} & 0.76 & \colorbox{wrongPredictionColor}{0.67} & 1.\\
$\datasetFactcheckShuffled$ & $\modelPhiThreeFiveInstructGenVBLLMLPrFifty$ & $\estimatorNoReject$ & \colorbox{wrongPredictionColor}{0.44} & 1. & \colorbox{correctPredictionColor}{\allRejected} & 0. & \colorbox{correctPredictionColor}{1.} & 0.44 & \colorbox{wrongPredictionColor}{0.} & 0.56 & \colorbox{wrongPredictionColor}{0.44} & 1.\\

$\datasetFactcheck$ $\calibrationSplit$ & $\modelPhiThreeFiveInstructGenVBLLMLPrFifty$ & $\estimatorVBLL$ & \colorbox{correctPredictionColor}{0.98} & 0.33 & \colorbox{correctPredictionColor}{0.99} & 0.29 & \colorbox{correctPredictionColor}{0.99} & 0.33 & \colorbox{correctPredictionColor}{0.98} & 0.29 & \colorbox{correctPredictionColor}{0.99} & 0.62\\
$\datasetFactcheck$ & $\modelPhiThreeFiveInstructGenVBLLMLPrFifty$ & $\estimatorVBLL$ & \colorbox{wrongPredictionColor}{0.29} & 0.06 & \colorbox{correctPredictionColor}{0.98} & 0.24 & \colorbox{wrongPredictionColor}{0.80} & 0.02 & \colorbox{wrongPredictionColor}{0.86} & 0.28 & \colorbox{wrongPredictionColor}{0.85} & 0.30\\
$\datasetFactcheckShuffled$ & $\modelPhiThreeFiveInstructGenVBLLMLPrFifty$ & $\estimatorVBLL$ & \colorbox{wrongPredictionColor}{0.22} & 0.21 & \colorbox{correctPredictionColor}{\allRejected} & 0. & \colorbox{correctPredictionColor}{1.} & 0.04 & \colorbox{wrongPredictionColor}{0.} & 0.16 & \colorbox{wrongPredictionColor}{0.22} & 0.21\\

\midrule

$\datasetFactcheck$ $\calibrationSplit$ & $\modelMixtralDiscVBLLMLP$ & $\estimatorNoReject$ & \colorbox{wrongPredictionColor}{0.91} & 0.50 & \colorbox{wrongPredictionColor}{0.93} & 0.50 & \colorbox{wrongPredictionColor}{0.93} & 0.49 & \colorbox{wrongPredictionColor}{0.91} & 0.51 & \colorbox{wrongPredictionColor}{0.92} & 1.\\
$\datasetFactcheck$ & $\modelMixtralDiscVBLLMLP$ & $\estimatorNoReject$ & \colorbox{wrongPredictionColor}{0.66} & 0.51 & \colorbox{wrongPredictionColor}{0.88} & 0.49 & \colorbox{wrongPredictionColor}{0.86} & 0.40 & \colorbox{wrongPredictionColor}{0.71} & 0.60 & \colorbox{wrongPredictionColor}{0.77} & 1.\\
$\datasetFactcheckShuffled$ & $\modelMixtralDiscVBLLMLP$ & $\estimatorNoReject$ & \colorbox{wrongPredictionColor}{0.84} & 1. & \colorbox{correctPredictionColor}{\allRejected} & 0. & \colorbox{correctPredictionColor}{1.} & 0.84 & \colorbox{wrongPredictionColor}{0.} & 0.16 & \colorbox{wrongPredictionColor}{0.84} & 1.\\

$\datasetFactcheck$ $\calibrationSplit$ & $\modelMixtralDiscVBLLMLP$ & $\estimatorVBLL$ & \colorbox{correctPredictionColor}{0.98} & 0.32 & \colorbox{correctPredictionColor}{0.99} & 0.31 & \colorbox{correctPredictionColor}{0.99} & 0.31 & \colorbox{correctPredictionColor}{0.98} & 0.31 & \colorbox{correctPredictionColor}{0.99} & 0.62\\
$\datasetFactcheck$ & $\modelMixtralDiscVBLLMLP$ & $\estimatorVBLL$ & \colorbox{wrongPredictionColor}{0.85} & 0.08 & \colorbox{correctPredictionColor}{0.97} & 0.27 & \colorbox{wrongPredictionColor}{0.89} & 0.08 & \colorbox{correctPredictionColor}{0.95} & 0.27 & \colorbox{wrongPredictionColor}{0.94} & 0.35\\
$\datasetFactcheckShuffled$ & $\modelMixtralDiscVBLLMLP$ & $\estimatorVBLL$ & \colorbox{wrongPredictionColor}{0.91} & 0.14 & \colorbox{correctPredictionColor}{\allRejected} & 0. & \colorbox{correctPredictionColor}{1.} & 0.13 & \colorbox{wrongPredictionColor}{0.} & 0.01 & \colorbox{wrongPredictionColor}{0.91} & 0.14\\

\midrule

$\datasetFactcheck$ $\calibrationSplit$ & $\modelMixtralDiscVBLLMLPrFifty$ & $\estimatorNoReject$ & \colorbox{wrongPredictionColor}{0.90} & 0.50 & \colorbox{wrongPredictionColor}{0.94} & 0.50 & \colorbox{wrongPredictionColor}{0.94} & 0.48 & \colorbox{wrongPredictionColor}{0.90} & 0.52 & \colorbox{wrongPredictionColor}{0.92} & 1.\\
$\datasetFactcheck$ & $\modelMixtralDiscVBLLMLPrFifty$ & $\estimatorNoReject$ & \colorbox{wrongPredictionColor}{0.62} & 0.51 & \colorbox{wrongPredictionColor}{0.89} & 0.49 & \colorbox{wrongPredictionColor}{0.86} & 0.37 & \colorbox{wrongPredictionColor}{0.69} & 0.63 & \colorbox{wrongPredictionColor}{0.75} & 1.\\
$\datasetFactcheckShuffled$ & $\modelMixtralDiscVBLLMLPrFifty$ & $\estimatorNoReject$ & \colorbox{wrongPredictionColor}{0.81} & 1. & \colorbox{correctPredictionColor}{\allRejected} & 0. & \colorbox{correctPredictionColor}{1.} & 0.81 & \colorbox{wrongPredictionColor}{0.} & 0.19 & \colorbox{wrongPredictionColor}{0.81} & 1.\\

$\datasetFactcheck$ $\calibrationSplit$ & $\modelMixtralDiscVBLLMLPrFifty$ & $\estimatorVBLL$ & \colorbox{correctPredictionColor}{0.98} & 0.33 & \colorbox{correctPredictionColor}{0.99} & 0.34 & \colorbox{correctPredictionColor}{0.99} & 0.32 & \colorbox{correctPredictionColor}{0.98} & 0.34 & \colorbox{correctPredictionColor}{0.99} & 0.67\\
$\datasetFactcheck$ & $\modelMixtralDiscVBLLMLPrFifty$ & $\estimatorVBLL$ & \colorbox{wrongPredictionColor}{0.72} & 0.12 & \colorbox{correctPredictionColor}{0.97} & 0.30 & \colorbox{wrongPredictionColor}{0.91} & 0.09 & \colorbox{wrongPredictionColor}{0.90} & 0.32 & \colorbox{wrongPredictionColor}{0.90} & 0.42\\
$\datasetFactcheckShuffled$ & $\modelMixtralDiscVBLLMLPrFifty$ & $\estimatorVBLL$ & \colorbox{wrongPredictionColor}{0.82} & 0.16 & \colorbox{correctPredictionColor}{\allRejected} & 0. & \colorbox{correctPredictionColor}{1.} & 0.13 & \colorbox{wrongPredictionColor}{0.} & 0.03 & \colorbox{wrongPredictionColor}{0.82} & 0.16\\

\midrule

$\datasetFactcheck$ $\calibrationSplit$ & $\modelMixtralGenVBLLMLP$ & $\estimatorNoReject$ & \colorbox{wrongPredictionColor}{0.91} & 0.48 & \colorbox{wrongPredictionColor}{0.93} & 0.52 & \colorbox{wrongPredictionColor}{0.92} & 0.48 & \colorbox{wrongPredictionColor}{0.92} & 0.52 & \colorbox{wrongPredictionColor}{0.92} & 1.\\
$\datasetFactcheck$ & $\modelMixtralGenVBLLMLP$ & $\estimatorNoReject$ & \colorbox{wrongPredictionColor}{0.63} & 0.51 & \colorbox{wrongPredictionColor}{0.88} & 0.49 & \colorbox{wrongPredictionColor}{0.85} & 0.38 & \colorbox{wrongPredictionColor}{0.70} & 0.62 & \colorbox{wrongPredictionColor}{0.76} & 1.\\
$\datasetFactcheckShuffled$ & $\modelMixtralGenVBLLMLP$ & $\estimatorNoReject$ & \colorbox{wrongPredictionColor}{0.73} & 1. & \colorbox{correctPredictionColor}{\allRejected} & 0. & \colorbox{correctPredictionColor}{1.} & 0.73 & \colorbox{wrongPredictionColor}{0.} & 0.27 & \colorbox{wrongPredictionColor}{0.73} & 1.\\

$\datasetFactcheck$ $\calibrationSplit$ & $\modelMixtralGenVBLLMLP$ & $\estimatorVBLL$ & \colorbox{correctPredictionColor}{0.96} & 0.32 & \colorbox{correctPredictionColor}{0.99} & 0.40 & \colorbox{correctPredictionColor}{0.99} & 0.31 & \colorbox{correctPredictionColor}{0.97} & 0.41 & \colorbox{correctPredictionColor}{0.98} & 0.72\\
$\datasetFactcheck$ & $\modelMixtralGenVBLLMLP$ & $\estimatorVBLL$ & \colorbox{wrongPredictionColor}{0.58} & 0.13 & \colorbox{correctPredictionColor}{0.99} & 0.33 & \colorbox{correctPredictionColor}{0.95} & 0.08 & \colorbox{wrongPredictionColor}{0.85} & 0.38 & \colorbox{wrongPredictionColor}{0.87} & 0.47\\
$\datasetFactcheckShuffled$ & $\modelMixtralGenVBLLMLP$ & $\estimatorVBLL$ & \colorbox{wrongPredictionColor}{0.54} & 0.11 & \colorbox{correctPredictionColor}{\allRejected} & 0. & \colorbox{correctPredictionColor}{1.} & 0.06 & \colorbox{wrongPredictionColor}{0.} & 0.05 & \colorbox{wrongPredictionColor}{0.54} & 0.11\\

\midrule

$\datasetFactcheck$ $\calibrationSplit$ & $\modelMixtralGenVBLLMLPrFifty$ & $\estimatorNoReject$ & \colorbox{wrongPredictionColor}{0.92} & 0.48 & \colorbox{wrongPredictionColor}{0.91} & 0.52 & \colorbox{wrongPredictionColor}{0.90} & 0.49 & \colorbox{wrongPredictionColor}{0.93} & 0.51 & \colorbox{wrongPredictionColor}{0.91} & 1.\\
$\datasetFactcheck$ & $\modelMixtralGenVBLLMLPrFifty$ & $\estimatorNoReject$ & \colorbox{wrongPredictionColor}{0.67} & 0.51 & \colorbox{wrongPredictionColor}{0.89} & 0.49 & \colorbox{wrongPredictionColor}{0.87} & 0.40 & \colorbox{wrongPredictionColor}{0.72} & 0.60 & \colorbox{wrongPredictionColor}{0.78} & 1.\\
$\datasetFactcheckShuffled$ & $\modelMixtralGenVBLLMLPrFifty$ & $\estimatorNoReject$ & \colorbox{wrongPredictionColor}{0.69} & 1. & \colorbox{correctPredictionColor}{\allRejected} & 0. & \colorbox{correctPredictionColor}{1.} & 0.69 & \colorbox{wrongPredictionColor}{0.} & 0.31 & \colorbox{wrongPredictionColor}{0.69} & 1.\\

$\datasetFactcheck$ $\calibrationSplit$ & $\modelMixtralGenVBLLMLPrFifty$ & $\estimatorVBLL$ & \colorbox{correctPredictionColor}{0.97} & 0.38 & \colorbox{correctPredictionColor}{0.98} & 0.40 & \colorbox{correctPredictionColor}{0.98} & 0.37 & \colorbox{correctPredictionColor}{0.97} & 0.41 & \colorbox{correctPredictionColor}{0.97} & 0.78\\
$\datasetFactcheck$ & $\modelMixtralGenVBLLMLPrFifty$ & $\estimatorVBLL$ & \colorbox{wrongPredictionColor}{0.67} & 0.19 & \colorbox{correctPredictionColor}{0.95} & 0.35 & \colorbox{wrongPredictionColor}{0.89} & 0.14 & \colorbox{wrongPredictionColor}{0.84} & 0.39 & \colorbox{wrongPredictionColor}{0.85} & 0.53\\
$\datasetFactcheckShuffled$ & $\modelMixtralGenVBLLMLPrFifty$ & $\estimatorVBLL$ & \colorbox{wrongPredictionColor}{0.67} & 0.15 & \colorbox{correctPredictionColor}{\allRejected} & 0. & \colorbox{correctPredictionColor}{1.} & 0.10 & \colorbox{wrongPredictionColor}{0.} & 0.05 & \colorbox{wrongPredictionColor}{0.67} & 0.15\\

    \bottomrule
  \end{tabular}
  }  
    \caption{Comparison of Bayesian last-layer estimators \citep{HarrisonEtAl-2024-VBLL} for the factcheck datasets, including the shuffled challenge sets, with \colorbox{correctPredictionColor}{$\alpha$}$=0.95$. \colorbox{correctPredictionColor}{\allRejected} indicates all predictions were rejected, which is preferred over falling \colorbox{wrongPredictionColor}{under} the expected accuracy. 
  $n=|\text{Admitted}|$, the count of non-rejected documents.} %
  \label{tab:experiments-factcheck-additional} 
\end{table*}

%% file: sdm.bbl
\begin{thebibliography}{37}
\providecommand{\natexlab}[1]{#1}

\bibitem[{Abdin et~al.(2024)Abdin, Aneja, Awadalla, Awadallah, Awan, Bach,
  Bahree, Bakhtiari, Bao, Behl, Benhaim, Bilenko, Bjorck, Bubeck, Cai, Cai,
  Chaudhary, Chen, Chen, Chen, Chen, Chen, Cheng, Chopra, Dai, Dixon, Eldan,
  Fragoso, Gao, Gao, Gao, Garg, Giorno, Goswami, Gunasekar, Haider, Hao,
  Hewett, Hu, Huynh, Iter, Jacobs, Javaheripi, Jin, Karampatziakis, Kauffmann,
  Khademi, Kim, Kim, Kurilenko, Lee, Lee, Li, Li, Liang, Liden, Lin, Lin, Liu,
  Liu, Liu, Liu, Liu, Luo, Madan, Mahmoudzadeh, Majercak, Mazzola, Mendes,
  Mitra, Modi, Nguyen, Norick, Patra, Perez-Becker, Portet, Pryzant, Qin,
  Radmilac, Ren, de~Rosa, Rosset, Roy, Ruwase, Saarikivi, Saied, Salim,
  Santacroce, Shah, Shang, Sharma, Shen, Shukla, Song, Tanaka, Tupini,
  Vaddamanu, Wang, Wang, Wang, Wang, Wang, Wang, Ward, Wen, Witte, Wu, Wu,
  Wyatt, Xiao, Xu, Xu, Xu, Xue, Yadav, Yang, Yang, Yang, Yang, Yu, Yuan, Zhang,
  Zhang, Zhang, Zhang, Zhang, Zhang, Zhang, and
  Zhou}]{Abdin-2024-Phi3-TechReport}
Marah Abdin, Jyoti Aneja, Hany Awadalla, Ahmed Awadallah, Ammar~Ahmad Awan,
  Nguyen Bach, Amit Bahree, Arash Bakhtiari, Jianmin Bao, Harkirat Behl, Alon
  Benhaim, Misha Bilenko, Johan Bjorck, Sébastien Bubeck, Martin Cai, Qin Cai,
  Vishrav Chaudhary, Dong Chen, Dongdong Chen, and 110 others. 2024.
\newblock \href {https://arxiv.org/abs/2404.14219} {Phi-3 technical report: A
  highly capable language model locally on your phone}.
\newblock \emph{Preprint}, arXiv:2404.14219.

\bibitem[{Angelopoulos et~al.(2021)Angelopoulos, Bates, Jordan, and
  Malik}]{Angelopoulos-2021-RAPS}
Anastasios~Nikolas Angelopoulos, Stephen Bates, Michael Jordan, and Jitendra
  Malik. 2021.
\newblock \href {https://openreview.net/forum?id=eNdiU_DbM9} {{Uncertainty Sets
  for Image Classifiers using Conformal Prediction}}.
\newblock In \emph{International Conference on Learning Representations}.

\bibitem[{Azaria and Mitchell(2023)}]{azaria-mitchell-2023-internal}
Amos Azaria and Tom Mitchell. 2023.
\newblock \href {https://doi.org/10.18653/v1/2023.findings-emnlp.68} {The
  internal state of an {LLM} knows when it{'}s lying}.
\newblock In \emph{Findings of the Association for Computational Linguistics:
  EMNLP 2023}, pages 967--976, Singapore. Association for Computational
  Linguistics.

\bibitem[{Brier(1950)}]{Brier-1950-BrierCalibration}
Glenn~W. Brier. 1950.
\newblock \href
  {https://doi.org/10.1175/1520-0493(1950)078<0001:VOFEIT>2.0.CO;2}
  {Verification of forecasts expressed in terms of probability}.
\newblock \emph{Monthly Weather Review}, 78(1):1 -- 3.

\bibitem[{Chow(1957)}]{Chow-1957-EarlyPredictWithRejectSystem}
C.~K. Chow. 1957.
\newblock \href {https://doi.org/10.1109/TEC.1957.5222035} {An optimum
  character recognition system using decision functions}.
\newblock \emph{IRE Transactions on Electronic Computers}, EC-6(4):247--254.

\bibitem[{Clevert et~al.(2016)Clevert, Unterthiner, and
  Hochreiter}]{ClevertEtal-2016-ELU}
Djork{-}Arn{\'{e}} Clevert, Thomas Unterthiner, and Sepp Hochreiter. 2016.
\newblock \href {http://arxiv.org/abs/1511.07289} {Fast and accurate deep
  network learning by exponential linear units (elus)}.
\newblock In \emph{4th International Conference on Learning Representations,
  {ICLR} 2016, San Juan, Puerto Rico, May 2-4, 2016, Conference Track
  Proceedings}.

\bibitem[{Cover and Hart(1967)}]{CoverAndHart-1967-KNNBayesError}
T.~Cover and P.~Hart. 1967.
\newblock \href {https://doi.org/10.1109/TIT.1967.1053964} {Nearest neighbor
  pattern classification}.
\newblock \emph{IEEE Transactions on Information Theory}, 13(1):21--27.

\bibitem[{Dawid(1982)}]{Dawid-1982-CalibratedBayesian}
A.~P. Dawid. 1982.
\newblock \href {https://doi.org/10.1080/01621459.1982.10477856} {The
  well-calibrated bayesian}.
\newblock \emph{Journal of the American Statistical Association},
  77(379):605--610.

\bibitem[{Devroye et~al.(1996)Devroye, Gy{\"o}rfi, and
  Lugosi}]{DevroyeEtAl-1996-APT}
Luc Devroye, L{\'a}szl{\'o} Gy{\"o}rfi, and G{\'a}bor Lugosi. 1996.
\newblock {A Probabilistic Theory of Pattern Recognition}.
\newblock In \emph{Stochastic Modelling and Applied Probability}.

\bibitem[{Dvoretzky et~al.(1956)Dvoretzky, Kiefer, and
  Wolfowitz}]{DKW-1956-DKW-Inequality}
A.~Dvoretzky, J.~Kiefer, and J.~Wolfowitz. 1956.
\newblock \href {https://doi.org/10.1214/aoms/1177728174} {{Asymptotic Minimax
  Character of the Sample Distribution Function and of the Classical
  Multinomial Estimator}}.
\newblock \emph{The Annals of Mathematical Statistics}, 27(3):642 -- 669.

\bibitem[{Foygel~Barber et~al.(2020)Foygel~Barber, Cand{\`e}s, Ramdas, and
  Tibshirani}]{BarberEtAl-2020-LimitsOfDistributionFree}
Rina Foygel~Barber, Emmanuel~J Cand{\`e}s, Aaditya Ramdas, and Ryan~J
  Tibshirani. 2020.
\newblock \href {https://doi.org/10.1093/imaiai/iaaa017} {{The limits of
  distribution-free conditional predictive inference}}.
\newblock \emph{Information and Inference: A Journal of the IMA},
  10(2):455--482.

\bibitem[{Gal and Ghahramani(2016)}]{GalAndZoubin-2016-MCDropout}
Yarin Gal and Zoubin Ghahramani. 2016.
\newblock \href {https://proceedings.mlr.press/v48/gal16.html} {Dropout as a
  bayesian approximation: Representing model uncertainty in deep learning}.
\newblock In \emph{Proceedings of The 33rd International Conference on Machine
  Learning}, volume~48 of \emph{Proceedings of Machine Learning Research},
  pages 1050--1059, New York, New York, USA. PMLR.

\bibitem[{Geifman and
  El-Yaniv(2017)}]{GeifmanAndEl-Yaniv-2017-NN-SelectiveClassification}
Yonatan Geifman and Ran El-Yaniv. 2017.
\newblock \href
  {https://proceedings.neurips.cc/paper_files/paper/2017/file/4a8423d5e91fda00bb7e46540e2b0cf1-Paper.pdf}
  {Selective classification for deep neural networks}.
\newblock In \emph{Advances in Neural Information Processing Systems},
  volume~30. Curran Associates, Inc.

\bibitem[{Guo et~al.(2017)Guo, Pleiss, Sun, and
  Weinberger}]{GuoEtAl-2017-TempScaling}
Chuan Guo, Geoff Pleiss, Yu~Sun, and Kilian~Q. Weinberger. 2017.
\newblock \href {https://proceedings.mlr.press/v70/guo17a.html} {On calibration
  of modern neural networks}.
\newblock In \emph{Proceedings of the 34th International Conference on Machine
  Learning}, volume~70 of \emph{Proceedings of Machine Learning Research},
  pages 1321--1330. PMLR.

\bibitem[{Gupta and Ramdas(2022)}]{GuptaAndRamdas-2022-ToplabelCalibration}
Chirag Gupta and Aaditya Ramdas. 2022.
\newblock \href {https://openreview.net/forum?id=WqoBaaPHS-} {{Top-label
  calibration and multiclass-to-binary reductions}}.
\newblock In \emph{International Conference on Learning Representations}.

\bibitem[{Harrison et~al.(2024)Harrison, Willes, and
  Snoek}]{HarrisonEtAl-2024-VBLL}
James Harrison, John Willes, and Jasper Snoek. 2024.
\newblock \href {https://openreview.net/forum?id=Sx7BIiPzys} {Variational
  bayesian last layers}.
\newblock In \emph{The Twelfth International Conference on Learning
  Representations}.

\bibitem[{Hwang and Ding(1997)}]{hwang-and-ding-1997}
J.~T.~Gene Hwang and A.~Adam Ding. 1997.
\newblock \href {http://www.jstor.org/stable/2965723} {Prediction intervals for
  artificial neural networks}.
\newblock \emph{Journal of the American Statistical Association},
  92(438):748--757.

\bibitem[{Jiang et~al.(2024)Jiang, Sablayrolles, Roux, Mensch, Savary, Bamford,
  Chaplot, de~las Casas, Hanna, Bressand, Lengyel, Bour, Lample, Lavaud,
  Saulnier, Lachaux, Stock, Subramanian, Yang, Antoniak, Scao, Gervet, Lavril,
  Wang, Lacroix, and Sayed}]{JiangEtAl-2024-Mixtral8x7B}
Albert~Q. Jiang, Alexandre Sablayrolles, Antoine Roux, Arthur Mensch, Blanche
  Savary, Chris Bamford, Devendra~Singh Chaplot, Diego de~las Casas, Emma~Bou
  Hanna, Florian Bressand, Gianna Lengyel, Guillaume Bour, Guillaume Lample,
  Lélio~Renard Lavaud, Lucile Saulnier, Marie-Anne Lachaux, Pierre Stock,
  Sandeep Subramanian, Sophia Yang, and 7 others. 2024.
\newblock \href {https://arxiv.org/abs/2401.04088} {Mixtral of experts}.
\newblock \emph{Preprint}, arXiv:2401.04088.

\bibitem[{Kingma and Ba(2015)}]{Kingma-2017-Adam-Optimizer}
Diederik~P. Kingma and Jimmy Ba. 2015.
\newblock \href {http://arxiv.org/abs/1412.6980} {Adam: {A} method for
  stochastic optimization}.
\newblock In \emph{3rd International Conference on Learning Representations,
  {ICLR} 2015, San Diego, CA, USA, May 7-9, 2015, Conference Track
  Proceedings}.

\bibitem[{Kull et~al.(2019)Kull, Perello-Nieto, K\"{a}ngsepp, Silva~Filho,
  Song, and Flach}]{KullEtAl-2019-BeyondTempScaling}
Meelis Kull, Miquel Perello-Nieto, Markus K\"{a}ngsepp, Telmo Silva~Filho, Hao
  Song, and Peter Flach. 2019.
\newblock \href
  {https://proceedings.neurips.cc/paper_files/paper/2019/file/8ca01ea920679a0fe3728441494041b9-Paper.pdf}
  {Beyond temperature scaling: Obtaining well-calibrated multiclass
  probabilities with dirichlet calibration}.
\newblock In \emph{Advances in Neural Information Processing Systems},
  volume~32. Curran Associates, Inc.

\bibitem[{Lakshminarayanan et~al.(2017)Lakshminarayanan, Pritzel, and
  Blundell}]{Lakshminarayanan-2017-DeepEnsembles}
Balaji Lakshminarayanan, Alexander Pritzel, and Charles Blundell. 2017.
\newblock \href
  {https://proceedings.neurips.cc/paper_files/paper/2017/file/9ef2ed4b7fd2c810847ffa5fa85bce38-Paper.pdf}
  {Simple and scalable predictive uncertainty estimation using deep ensembles}.
\newblock In \emph{Advances in Neural Information Processing Systems},
  volume~30. Curran Associates, Inc.

\bibitem[{Lei and Wasserman(2014)}]{LeiAndWasserman-2014-PredictionBands}
Jing Lei and Larry Wasserman. 2014.
\newblock \href {https://doi.org/10.1111/rssb.12021} {Distribution-free
  prediction bands for non-parametric regression}.
\newblock \emph{Journal of the Royal Statistical Society: Series B (Statistical
  Methodology)}, 76(1):71--96.

\bibitem[{Loshchilov and Hutter(2019)}]{LoshchilovAndHutter-2019-AdamW}
Ilya Loshchilov and Frank Hutter. 2019.
\newblock \href {https://openreview.net/forum?id=Bkg6RiCqY7} {Decoupled weight
  decay regularization}.
\newblock In \emph{International Conference on Learning Representations}.

\bibitem[{Maas et~al.(2011)Maas, Daly, Pham, Huang, Ng, and
  Potts}]{Maas-EtAl-2011-OriginalCitedSourceForIMDbReviewsData}
Andrew~L. Maas, Raymond~E. Daly, Peter~T. Pham, Dan Huang, Andrew~Y. Ng, and
  Christopher Potts. 2011.
\newblock \href {http://www.aclweb.org/anthology/P11-1015} {Learning word
  vectors for sentiment analysis}.
\newblock In \emph{Proceedings of the 49th Annual Meeting of the Association
  for Computational Linguistics: Human Language Technologies}, pages 142--150,
  Portland, Oregon, USA. Association for Computational Linguistics.

\bibitem[{Massart(1990)}]{Massart-1990-DKW-Tight-Constant}
P.~Massart. 1990.
\newblock \href {https://doi.org/10.1214/aop/1176990746} {{The Tight Constant
  in the Dvoretzky-Kiefer-Wolfowitz Inequality}}.
\newblock \emph{The Annals of Probability}, 18(3):1269 -- 1283.

\bibitem[{Ovadia et~al.(2019)Ovadia, Fertig, Ren, Nado, Sculley, Nowozin,
  Dillon, Lakshminarayanan, and Snoek}]{Ovadia-EtAl-2019-EvaluatingUncertainty}
Yaniv Ovadia, Emily Fertig, Jie Ren, Zachary Nado, D.~Sculley, Sebastian
  Nowozin, Joshua Dillon, Balaji Lakshminarayanan, and Jasper Snoek. 2019.
\newblock \href
  {https://proceedings.neurips.cc/paper_files/paper/2019/file/8558cb408c1d76621371888657d2eb1d-Paper.pdf}
  {Can you trust your model\textquotesingle s uncertainty? evaluating
  predictive uncertainty under dataset shift}.
\newblock In \emph{Advances in Neural Information Processing Systems},
  volume~32. Curran Associates, Inc.

\bibitem[{Platt(1999)}]{Platt-1999-PlattScaling}
John~C. Platt. 1999.
\newblock {Probabilistic Outputs for Support Vector Machines and Comparisons to
  Regularized Likelihood Methods}.
\newblock In \emph{Advances in Large Margin Classifiers}, pages 61--74. MIT
  Press.

\bibitem[{Romano et~al.(2020)Romano, Sesia, and Candes}]{RomanoEtAl-2020-APS}
Yaniv Romano, Matteo Sesia, and Emmanuel Candes. 2020.
\newblock \href
  {https://proceedings.neurips.cc/paper_files/paper/2020/file/244edd7e85dc81602b7615cd705545f5-Paper.pdf}
  {Classification with valid and adaptive coverage}.
\newblock In \emph{Advances in Neural Information Processing Systems},
  volume~33, pages 3581--3591. Curran Associates, Inc.

\bibitem[{Rosenthal et~al.(2017)Rosenthal, Farra, and
  Nakov}]{Rosenthal-etal-2017-Semeval-Task4}
Sara Rosenthal, Noura Farra, and Preslav Nakov. 2017.
\newblock \href {https://doi.org/10.18653/v1/S17-2088} {{S}em{E}val-2017 task
  4: Sentiment analysis in {T}witter}.
\newblock In \emph{Proceedings of the 11th International Workshop on Semantic
  Evaluation ({S}em{E}val-2017)}, pages 502--518, Vancouver, Canada.
  Association for Computational Linguistics.

\bibitem[{Schmaltz(2021)}]{schmaltz-2021-insights}
Allen Schmaltz. 2021.
\newblock \href {https://doi.org/10.1162/coli_a_00416} {Detecting local
  insights from global labels: Supervised and zero-shot sequence labeling via a
  convolutional decomposition}.
\newblock \emph{Computational Linguistics}, 47(4):729--773.

\bibitem[{Shazeer et~al.(2017)Shazeer, Mirhoseini, Maziarz, Davis, Le, Hinton,
  and Dean}]{ShazeerEtAl-2017-MOE}
Noam Shazeer, *Azalia Mirhoseini, *Krzysztof Maziarz, Andy Davis, Quoc Le,
  Geoffrey Hinton, and Jeff Dean. 2017.
\newblock \href {https://openreview.net/forum?id=B1ckMDqlg} {Outrageously large
  neural networks: The sparsely-gated mixture-of-experts layer}.
\newblock In \emph{International Conference on Learning Representations}.

\bibitem[{Vaicenavicius et~al.(2019)Vaicenavicius, Widmann, Andersson,
  Lindsten, Roll, and Sch\"{o}n}]{VaicenaviciusEtAl-2019-EvaluatingCalibration}
Juozas Vaicenavicius, David Widmann, Carl Andersson, Fredrik Lindsten, Jacob
  Roll, and Thomas Sch\"{o}n. 2019.
\newblock \href {https://proceedings.mlr.press/v89/vaicenavicius19a.html}
  {Evaluating model calibration in classification}.
\newblock In \emph{Proceedings of the Twenty-Second International Conference on
  Artificial Intelligence and Statistics}, volume~89 of \emph{Proceedings of
  Machine Learning Research}, pages 3459--3467. PMLR.

\bibitem[{Valiant(1984)}]{Valiant-1984-PAC}
L.~G. Valiant. 1984.
\newblock \href {https://doi.org/10.1145/1968.1972} {A theory of the
  learnable}.
\newblock \emph{Commun. ACM}, 27(11):1134–1142.

\bibitem[{Vaswani et~al.(2017)Vaswani, Shazeer, Parmar, Uszkoreit, Jones,
  Gomez, Kaiser, and Polosukhin}]{VaswaniEtAl-2017-AttentionIsAllYouNeed}
Ashish Vaswani, Noam Shazeer, Niki Parmar, Jakob Uszkoreit, Llion Jones,
  Aidan~N Gomez, {\L}ukasz Kaiser, and Illia Polosukhin. 2017.
\newblock \href
  {https://proceedings.neurips.cc/paper_files/paper/2017/file/3f5ee243547dee91fbd053c1c4a845aa-Paper.pdf}
  {Attention is all you need}.
\newblock In \emph{Advances in Neural Information Processing Systems},
  volume~30. Curran Associates, Inc.

\bibitem[{Vovk(2012)}]{Vovk-2012-ConditionalValidity}
Vladimir Vovk. 2012.
\newblock \href {https://proceedings.mlr.press/v25/vovk12.html} {Conditional
  validity of inductive conformal predictors}.
\newblock In \emph{Proceedings of the Asian Conference on Machine Learning},
  volume~25 of \emph{Proceedings of Machine Learning Research}, pages 475--490,
  Singapore Management University, Singapore. PMLR.

\bibitem[{Vovk et~al.(2005)Vovk, Gammerman, and
  Shafer}]{Vovk-2005-AlgorithmicLearningBook}
Vladimir Vovk, Alex Gammerman, and Glenn Shafer. 2005.
\newblock \emph{Algorithmic Learning in a Random World}.
\newblock Springer-Verlag, Berlin, Heidelberg.

\bibitem[{Zhang et~al.(2015)Zhang, Zhao, and
  LeCun}]{ZhangEtAl-CharacterCNNforClassification}
Xiang Zhang, Junbo Zhao, and Yann LeCun. 2015.
\newblock \href
  {https://proceedings.neurips.cc/paper_files/paper/2015/file/250cf8b51c773f3f8dc8b4be867a9a02-Paper.pdf}
  {Character-level convolutional networks for text classification}.
\newblock In \emph{Advances in Neural Information Processing Systems},
  volume~28. Curran Associates, Inc.

\end{thebibliography}
